\documentclass[a4paper]{article}
\usepackage{enumerate}
\usepackage{amsmath}
\usepackage{amssymb}
\usepackage{amsthm}
\usepackage{graphicx}
\usepackage{subfigure}
\usepackage{times}
\usepackage{color}
\usepackage{multirow}
\usepackage{bm}
\usepackage{xr}
\usepackage{url}
\usepackage{natbib}
\usepackage{centernot}

\usepackage{epstopdf}

\newcommand{\parder}[1]{\frac{\partial}{\partial #1}}
\newtheorem{theorem}{Theorem}
\newtheorem{proposition}{Proposition}
\newcommand{\intd}{\mathrm{d}}
\newcommand{\dx}{d_{\mathrm{x}}}

\newcommand{\calD}{\mathcal{D}} 
\newcommand{\pst}{p^{\star}}
\newcommand{\qst}{q^{\star}}
\newcommand{\psta}[1]{p^{\star #1}}
\newcommand{\qsta}[1]{q^{\star #1}}
\newcommand{\epu}{\epsilon(\bm{u})} 
\usepackage{color}

\newcommand{\bigCI}{\mathrel{\text{\scalebox{1.07}{$\perp\mkern-10mu\perp$}}}}
\newcommand{\nbigCI}{\centernot{\bigCI}}
\newtheorem{lemma}{Lemma}

\newcommand{\vtheta}{\bm{\theta}}
\newcommand{\bx}{\bar{\bm{x}}}
\newcommand{\bu}{\bar{\bm{u}}}
\newcommand{\tf}{r_{\vtheta}(\bm{x},\bm{u})^{(\gamma+1)}}
\newcommand{\vhtheta}{\bm{\hat{\theta}}}
\newcommand{\intdx}{\intd\bm{x}}
\newcommand{\intdu}{\intd\bm{u}}
\usepackage[utf8]{inputenc}

\makeatletter
\newcommand\footnoteref[1]{\protected@xdef\@thefnmark{\ref{#1}}\@footnotemark}
\makeatother
\title{Robust contrastive learning\\ and nonlinear ICA in the
 presence of outliers}
\author{Hiroaki Sasaki\\
Department of Complex and Intelligent Systems\\
Future University Hakodate, Japan
\\\vspace{-1mm} \\
Takashi Takenouchi  \\
Department of Complex and Intelligent Systems\\
Future University Hakodate, Japan\\
RIKEN AIP, Japan
\\\vspace{-1mm} \\
Ricardo Monti \\ 
Gatsby Computational Neuroscience Unit \\
University College London, UK
\\\vspace{-1mm} \\
Aapo Hyv{\"a}rinen \\ 
Universit{\'e} Paris-Saclay
Inria, CEA, France\\
Department of Computer
Science \\ Helsinki Institute for Information Technology HIIT\\
University of Helsinki, Finland}
\date{}
\begin{document}
 \maketitle
 \begin{abstract}  
  Nonlinear independent component analysis (ICA) is a general framework
  for unsupervised representation learning, and aimed at recovering the
  latent variables in data. Recent practical methods perform nonlinear
  ICA by solving a series of classification problems based on logistic
  regression. However, it is well-known that logistic regression is
  vulnerable to outliers, and thus the performance can be strongly
  weakened by outliers. In this paper, we first theoretically analyze
  nonlinear ICA models in the presence of outliers. Our analysis implies
  that estimation in nonlinear ICA can be seriously hampered when
  outliers exist on the tails of the (noncontaminated) target density,
  which happens in a typical case of contamination by outliers.  We
  develop two robust nonlinear ICA methods based on the
  $\gamma$-divergence, which is a robust alternative to the
  KL-divergence in logistic regression.  The proposed methods are shown
  to have desired robustness properties in the context of nonlinear
  ICA. We also experimentally demonstrate that the proposed methods are
  very robust and outperform existing methods in the presence of
  outliers. Finally, the proposed method is applied to ICA-based causal
  discovery and shown to find a plausible causal relationship on fMRI
  data.
 \end{abstract} 
 \section{Introduction}
 \label{sec:intro}
 Nonlinear independent component analysis (ICA) is a principled
 framework for unsupervised representation learning which has generated
 a large amount of recent interest in learning deep neural
 networks. Unlike most unsupervised methods, nonlinear ICA is based on a
 clear statistical estimation task. The problem is rigorously formulated
 by defining a generative model for the data, and the goal is to recover
 (or identify) the mutually independent latent source components of
 which the data is observed as a general nonlinear mixing. Nonlinear ICA
 includes a number of potential applications such as causal
 analysis~\citep{monti2019causal} and transfer
 learning~\citep{noroozi2016unsupervised}.
 
 In contrast to the great success of \emph{linear}
 ICA~\citep{hyvarinen2000independent}, nonlinear ICA has not received so
 much attention until recently because the problem is fundamentally
 ill-posed: There are an infinite number of decompositions of a random
 vector into mutually independent
 variables~\citep{hyvarinen1999nonlinear,pmlr-v97-locatello19a}, while
 the identifiability proof is established in linear
 ICA~\citep{comon1994independent}. Thus, in general, we cannot recover
 the original source components under the same conditions as linear ICA.
 
 Novel identifiability proofs for nonlinear ICA have been recently
 established~\citep{sprekeler2014extension,
 hyvarinen2016unsupervised,pmlr-v54-hyvarinen17a,hyvarinen2018nonlinear}.
 The main idea is to introduce some auxiliary variables given which the
 latent source components are conditionally independent.  For instance,
 \emph{time contrastive learning} divides time series data into a number
 of time segments and uses the time segment label as the auxiliary
 variable~\citep{hyvarinen2016unsupervised}; in \emph{permutation
 contrastive learning}, the auxiliary variable is the history of
 time-series data~\citep{pmlr-v54-hyvarinen17a}.  These methods perform
 nonlinear ICA by solving a series of classification problems in a
 similar spirit as generative adversarial
 network~\citep{goodfellow2014generative} and noise contrastive
 estimation~\citep{Gutmann2012a}. Interestingly, a heuristic yet
 successful approach called \emph{self-supervised
 learning}~\citep{larsson2017colorization,noroozi2016unsupervised,oord2018representation}
 also takes the same approach of solving unsupervised learning problems
 through classification. Thus, the theory of nonlinear ICA might shed
 light on the principles underlying self-supervised learning.
 
 In order to solve the nonlinear ICA problems in practice, logistic
 regression has been
 employed~\citep{hyvarinen2016unsupervised,pmlr-v54-hyvarinen17a,hyvarinen2018nonlinear},
 which is based on (conditional) maximum likelihood estimation
 (MLE). MLE has a number of useful properties, but it is well-known to
 be vulnerable to outliers. Thus, the performance of the existing
 nonlinear ICA methods might be strongly degraded by outliers. This is a
 very important problem because outliers are ubiquitous on real-world
 datasets. For instance, outliers are often observed in functional MRI
 data to which ICA methods have been applied~\citep{monti2019causal}.
 
 In this paper, we first define a contaminated density model of sources
 as a mixture of the (noncontaminated) target and outlier densities, and
 then theoretically analyze how outliers hamper estimation in nonlinear
 ICA. Our analysis implies that estimation in nonlinear ICA may be
 degraded particularly when the ratio of the outlier density to the
 target density can take a very large value. This large ratio happens
 when the outlier density lies on the tails of the target density as in
 a common outlier situation.
 
 Next, we propose two robust methods for nonlinear ICA. Our methods also
 solve classification problems, but are based on the
 $\gamma$-divergence~\citep{fujisawa2008robust}. $\gamma$-divergence is
 a generalization of KL-divergence and has a favorable robust
 property~\citep{fujisawa2008robust,hung2018robust,kawashima2018difference},
 which is expressed as the \emph{super
 robustness}~\citep{cichocki2010families,amari2016information}: The
 latent bias caused from outliers can be sufficiently small even in the
 case of heavy contamination almost as if outliers did not exist. This
 is in stark contrast with the density power
 divergence~\citep{basu1998robust}, which is often theoretically proved
 to be robust under small contamination of outliers.  We show that the
 $\gamma$-divergence has a desirable robust property in the context of
 nonlinear ICA as well, and experimentally confirm that the proposed
 nonlinear ICA methods are much more robust against outliers than
 existing methods. Finally, our robust nonlinear ICA method is applied
 to causal analysis and demonstrated to find a plausible causal
 relationship on fMRI data.
 \section{Background} 
 \label{sec:background}
 ICA is a rigorous framework for unsupervised learning, and assumes that
 the $\dx$-dimensional vectors of observed data
 $\bm{x}(t):=(x_1(t),\dots,x_{\dx}(t))^{\top},~t=1,\dots,T$ are
 generated from a nonlinear mixing of the source vectors
 $\bm{s}(t)=(s_1(t),\dots, s_{\dx}(t))^{\top}$ as
 \begin{align}
  \bm{x}(t)=\bm{f}(\bm{s}(t)), \label{ICA-model}
 \end{align}
 where $\bm{f}(\bm{s})=(f_1(\bm{s}), \dots, f_{\dx}(\bm{s}))^{\top}$,
 and $f_i$ denotes a smooth and invertible nonlinear function. The goal
 is to recover (or identify) the sources from data only.
 
 Nonlinear ICA has been hampered by the fact that the problem is
 seriously ill-posed and the original sources cannot be recovered (i.e.,
 not identifiable) under the same independence assumption as linear ICA,
 although there were heuristic works
 previously~\citep{wiskott2002slow,harmeling2003kernel}. Recently, novel
 identifiability proofs have been established together with practical
 algorithms~\citep{sprekeler2014extension,
 hyvarinen2016unsupervised,pmlr-v54-hyvarinen17a,hyvarinen2018nonlinear}.
 Time contrastive learning (TCL) divides time series data
 $\{\bm{x}(t)\}_{t=1}^T$ into $K$ time segments, and then a time segment
 label $u(t)=k,~k=1,\dots, K$ is assigned to $\bm{x}(t)$ in the $k$-th
 segment. A nonlinear feature
 $\bm{h}(\bm{x}(t))=(h_1(\bm{x}(t)),\dots,h_{\dx}(\bm{x}(t)))^{\top}$
 modelled by a neural network is learned via multinomial logistic
 regression to the \emph{artificial} supervised dataset
 $\{(u(t),\bm{x}(t))\}_{t=1}^{T}$.  For the identifiability, when the
 conditional density of $\bm{s}$ given a time segment label $u$ is
 conditionally independent and belongs to an exponential family as,
 \begin{align}
  \log \pst(\bm{s}|u)=\sum_{j=1}^d \lambda_{u,j}\qst_j(s_j)
  +\lambda_{u,0}-\log Z(\bm{\lambda}_{u}), \label{exp-family-TCL}
 \end{align}
 where $\qst_j$ is a scalar function, $\lambda_{u,j}$ denotes a
 parameter depending on $u$,
 $\bm{\lambda}_{u}:=(\lambda_{y,0},\lambda_{u,1},\dots,\lambda_{u,\dx})^{\top}$,
 and $Z(\bm{\lambda}_{y})$ is the partition function, then Theorem~1
 in~\citet{hyvarinen2016unsupervised} states that the learned
 $\bm{h}(\bm{x})$ asymptotically equals to
 $\bm{q}(\bm{s})=(q_1(s_1),\dots,q_{\dx}(s_{\dx}))^{\top}$ up to a
 linear transformation.

 A more general theory without the exponential family
 assumption~\eqref{exp-family-TCL} was established
 in~\citet{hyvarinen2018nonlinear}. Suppose that some auxiliary data
 $\bm{u}(t)$ is available in addition to $\bm{x}(t)$. For instance, the
 time segment label in TCL can be interpreted as auxiliary data, and
 permutation contrastive learning employs the past information of
 $\bm{x}(t)$ (e.g,
 $\bm{u}(t)=\bm{x}(t-1)$)~\citep{pmlr-v54-hyvarinen17a}. In order to
 learn a nonlinear feature $\bm{h}(\bm{x})$, the following binary
 classification problem is solved by logistic regression:
 \begin{align}
  \hspace{-2mm}\calD:=\{(\bm{x}(t), \bm{u}(t)\}_{t=1}^{T}~\text{vs.}
  ~\calD_{\mathrm{p}}:=\{(\bm{x}(t), \bm{u}_{\mathrm{p}}(t))\}_{t=1}^{T},
  \label{binary-classification}
 \end{align}
 where $\bm{u}_{\mathrm{p}}(t)$ is a random permutation of $\bm{u}(t)$
 with respect to $t$. Eq.\eqref{binary-classification} indicates that
 $\calD$ is drawn from the joint density of $\bm{x}(t)$ and $\bm{u}(t)$,
 while the underlying density of $\calD_{\mathrm{p}}$ can be regarded as
 the product of marginal densities of $\bm{x}(t)$ and $\bm{u}(t)$. Under
 the conditional independence assumption even more general than the
 exponential family~\eqref{exp-family-TCL},
 \begin{align}
  \log \pst(\bm{s}|\bm{u})=\sum_{i=1}^{\dx}\qst(s_i|\bm{u})
  -\log Z(\bm{u}),
  \label{general-conditional-independence}
 \end{align}
 where $Z(\bm{u})$ denotes the partition function and $\qst$ is a twice
 differential function, it was proved that the learned $\bm{h}$ equals
 to $\bm{s}$ up to an invertible function when $\pst(\bm{s}|\bm{u})$ is
 sufficiently diverse and
 complex~\citep[Theorem~1]{hyvarinen2018nonlinear}.\footnote{ The
 complexity and diversity of $\pst(\bm{s}|\bm{u})$ is expressed by the
 \emph{Assumption of Variability} in~\citet{hyvarinen2018nonlinear}.}
 However, the identifiability theory for the general case does not hold
 in the exponential family case~\eqref{exp-family-TCL} because the
 exponential family is too
 ``simple''~\citep[Theorem~2]{hyvarinen2018nonlinear}. Thus, both
 theories for the general and exponential cases complement each other.

 The nonlinear ICA methods above employ logistic regression to learn
 $\bm{h}(\bm{x})$, which is based on (conditional) maximum likelihood
 estimation (MLE). MLE has a number of useful properties such as
 asymptotic efficiency~\citep{wasserman2006all}; on the other hand, it
 is well-known that MLE can be vulnerable against outliers. Thus, these
 nonlinear ICA methods might be sensitive to outliers. Next, we first
 theoretically investigate how outliers hamper estimation in nonlinear
 ICA, and then propose robust practical methods.
 \section{Influence of outliers in nonlinear ICA}
 \label{sec:outliers}
 This section theoretically investigates the influence of outliers in
 nonlinear ICA.
  \subsection{Contaminated density model by outliers}
  Here, we first assume the following contaminated conditional density
  model of $\bm{s}$ given auxiliary variables $\bm{u}$:
  \begin{align}
   p(\bm{s}|\bm{u})=(1-\epu)\pst(\bm{s}|\bm{u})+\epu\delta(\bm{s}|\bm{u}),
   \label{source-contamination-model}
  \end{align}
  where $\epu$ is a contamination ratio in $[0,1)$.
  Eq.\eqref{source-contamination-model} means that the sources from
  $\pst(\bm{s}|\bm{u})$ are \emph{contaminated} by outliers generated
  from the outlier density $\delta(\bm{s}|\bm{u})$. We call
  $\pst(\bm{s}|\bm{u})$ the \emph{target} density in this paper because
  it generates the target sources which we want to recover from data
  $\bm{x}$.  Since $\epu$ can be dependent on $\bm{u}$, the contaminated
  density model~\eqref{source-contamination-model} is very general and
  called \emph{heterogeneous} contamination.  As theoretically shown
  later, our methods can accommodate heterogeneous contamination.
  \subsection{Influence of outliers in conditionally exponential case}  
  As in~\citet[Section~4.3]{hyvarinen2018nonlinear}, we first focus on
  the following conditionally independent and exponential family, which
  generalizes the exponential family~\eqref{exp-family-TCL} in TCL:
  \begin{align}
   \log \pst(\bm{s}|\bm{u})=\sum_{j=1}^{\dx} \lambda_{j}(\bm{u})q_j(s_j)
  +\lambda_{0}(\bm{u})-\log Z(\bm{\lambda}(\bm{u})), \label{exp-family}
  \end{align}
  where $\bm{\lambda}(\bm{u}):=(\lambda_{0}(\bm{u}),
  \lambda_{1}(\bm{u}),\dots,\lambda_{\dx}(\bm{u}))^{\top}$.  Then, we
  investigate how the outlier density $\delta(\bm{s}|\bm{u})$ hampers
  estimation in ICA, and we establish the following theorem:
  \begin{theorem}  
   \label{theo:identifiability} First, the following assumptions are
   made:
   \begin{enumerate}[({A}1)]
    \item Data $\bm{x}$ is generated from~\eqref{ICA-model} where
  	  $\bm{f}$ is invertible.

    \item $\pst(\bm{s}|\bm{u})$ is conditionally independent and belongs
	  to the exponential family~\eqref{exp-family}.
	  
    \item For all $\bm{s}$ and $\bm{u}$,
	  $\frac{\delta(\bm{s}|\bm{u})}{\pst(\bm{s}|{\bm{u}})}$ is
	  finite.
	  	  
    \item In the limit of infinite data, $p(\bm{x}|\bm{u})$ is
	  universally approximated as
	  \begin{align}
	   \log\frac{p(\bm{x}|\bm{u})}{c(\bm{x})e(\bm{u})}
	   =\bm{w}(\bm{u})^{\top}\bm{h}(\bm{x}),
	   \label{univ-approx-exp}
	  \end{align}
	  where
	  $\bm{w}(\bm{u}):=(w_1(\bm{u}),\dots,w_{\dx}(\bm{u}))^{\top}$ is
	  a vector-valued function, and $c$ and $e$ some scalar
	  functions.

    \item There exist $m+1$ points $\bm{u}_0,\bm{u}_1,\dots,\bm{u}_m$
	  such that the following $\dx$ by $\dx$ matrices are
	  invertible:
	  $\bar{\bm{\Lambda}}:=\sum_{i=1}^m\bar{\bm{\lambda}}(\bm{u}_i)
	  \bar{\bm{\lambda}}(\bm{u}_i)^{\top}$ and
	  $\sum_{i=1}^m\bar{\bm{w}}(\bm{u}_i)
	  \bar{\bm{\lambda}}(\bm{u}_i)^{\top}$, where
	  $\bar{\bm{\lambda}}(\bm{u}):=\bm{\lambda}(\bm{u})-\bm{\lambda}(\bm{u}_0)$
	  and $\bar{\bm{w}}(\bm{u}):=\bm{w}(\bm{u})-\bm{w}(\bm{u}_0)$.
   \end{enumerate}
   Then, regarding sufficiently small $\epu$ for all $\bm{u}$, in the
   limit of infinite data,
   \begin{align}
    \bm{q}(\bm{s})+\bm{r}(\bm{s})    
    &=\bm{A}\bm{h}(\bm{x})+\bm{\alpha},
    \label{outlier-indetifiability}
   \end{align}
   where $\bm{A}$ is a $\dx$ by $\dx$ invertible matrix, $\bm{\alpha}$
   is a $\dx$-dimensional vector, and with
   $\bar{\bm{\omega}}(\bm{u}):=\bar{\bm{\Lambda}}^{-1}\bar{\bm{\lambda}}(\bm{u})$
   $\bm{1}_{\dx}=(1,1,\dots,1)^{\top}$ and
   $\epsilon_{\max}:=\max_{i=0,1,\dots,m}\epsilon(\bm{u}_i)$,
   \begin{align}
    \bm{r}(\bm{s})&:=
    \sum_{i=1}^m\left\{\epsilon(\bm{u}_i)
    \frac{\delta(\bm{s}|\bm{u}_i)}{\pst(\bm{s}|\bm{u}_i)}
    -\epsilon(\bm{u}_0)\frac{\delta(\bm{s}|\bm{u}_0)}{\pst(\bm{s}|\bm{u}_0)}\right\}
    \bar{\bm{\omega}}(\bm{u}_i)+O(\epsilon_{\max}^2)
    \bm{1}_{\dx}.
    \label{defi-r-conditional}
   \end{align}
  \end{theorem}  
  The proof is deferred to Appendix~\ref{app:identifiability}.  First of
  all, in the case of no outliers, \eqref{outlier-indetifiability} is
  essentially the same identifiability result as Theorem~3
  in~\citet{hyvarinen2018nonlinear} as well as Theorem~1
  in~\citet{hyvarinen2016unsupervised} for TCL: $\epu=0$ leads to
  $\bm{r}(\bm{s})=\bm{0}$, and therefore $\bm{h}(\bm{x})$ equals to
  $\bm{q}(\bm{s})$ up to a linear transformation. This linear
  indeterminacy could be removed by some linear ICA method in
  postprocessing.
  
  Again, in no outlier case (i.e.,
  $p(\bm{x}|\bm{u})=\pst(\bm{x}|\bm{u})$), Assumption~(A4) can be
  written as
  \begin{align}
   \log\frac{\pst(\bm{x}|\bm{u})}{c(\bm{x})e(\bm{u})}
   =\bm{w}(\bm{u})^{\top}\bm{h}(\bm{x}).
   \label{univ-approx-exp-target}
  \end{align}  
  To satisfy~\eqref{univ-approx-exp-target},
  \citet{hyvarinen2018nonlinear} performs binary logistic regression
  where the log-odds ratio,
  $\log\frac{\pst(\bm{x},\bm{u})}{\pst(\bm{x})p(\bm{u})}$, is
  approximated by $\bm{w}(\bm{u})^{\top}\bm{h}(\bm{x})$ where
  $\pst(\bm{x})=\int \pst(\bm{x},\bm{u})\intd\bm{u}$.
  Eq.\eqref{univ-approx-exp-target} (or Assumption~(A4)) is a more
  general expression than the odds ratio: We do not necessarily need to
  accurately estimate the noncontaminated log-odds ratio, and it is
  sufficient to perform nonlinear ICA that the numerator is the
  conditional density $\pst(\bm{s}|\bm{u})$ or joint density
  $\pst(\bm{s},\bm{u})$ up to the product of nonzero scalar functions of
  $\bm{s}$ and $\bm{u}$. In fact, this is the key point for our robust
  method proposed in Section~\ref{ssec:binary}.  
  
  On the other hand, when $\epu\neq 0$,
  Theorem~\ref{theo:identifiability} indicates that estimation for the
  exponential family might be hampered by $\bm{r}(\bm{s})$. In
  particular, the elements in $\bm{r}(\bm{s})$ can be significantly
  nonzeros if the density ratio
  $\frac{\delta(\bm{s}|\bm{u})}{\pst(\bm{s}|\bm{u})}$
  in~\eqref{defi-r-conditional} is very large. The large ratio tends to
  happen when $\delta(\bm{s}|\bm{u})$ lies on the tails of
  $\pst(\bm{s}|\bm{u})$ (i.e., very small $\pst(\bm{s}|\bm{u})$, but
  large $\delta(\bm{s}|\bm{u})$). Thus, since logistic regression
  accurately estimates the log-odds ratio of the contaminated density,
  existing nonlinear ICA methods might be sensitive to
  outliers. Therefore, it would be desirable to develop a robust
  nonlinear ICA method.

  \subsection{Influence of outliers in non-exponential case}
  We performed a similar contamination analysis as
  Theorem~\ref{theo:identifiability} under the general (non-exponential)
  conditional independence
  condition~\eqref{general-conditional-independence} as well, but put
  the details in Appendix~\ref{app:general-case}. The conclusion is
  sightly complicated yet fundamentally similar as
  Theorem~\eqref{theo:identifiability}: Estimation in nonlinear ICA can
  be hampered when the four ratios,
  $\frac{\delta^{m}(\bm{s}|\bm{u})}{\pst(\bm{s}|\bm{u})}$,
  $\frac{\delta^{l}(\bm{s}|\bm{u})}{\pst(\bm{s}|\bm{u})}$,
  $\frac{\delta(\bm{s}|\bm{u})}{\pst(\bm{s}|\bm{u})}$ and
  $\frac{\delta^{l,m}(\bm{s}|\bm{u})}{\pst(\bm{s}|\bm{u})}$, are very
  large where $\delta^{l}(\bm{s}|\bm{u}):=\frac{\partial
  \delta(\bm{s}|\bm{u})}{\partial s_l}$ and
  $\delta^{l,m}(\bm{s}|\bm{u}):=\frac{\partial^2
  \delta(\bm{s}|\bm{u})}{\partial s_l\partial s_m}$ for
  $l,m=1,\dots,\dx$.  These ratios might be large when smooth
  $\delta(\bm{s}|\bm{u})$ exists on the tails of $\pst(\bm{s}|\bm{u})$.
  Therefore, again, it would be useful to develop a robust method in the
  general non-exponential case as well.
 \section{Robust contrastive learning}
 \label{sec:RCL}
 Our goal is to robustify nonlinear ICA methods. In light of the results
 above, the key is to estimate
 $\frac{\pst(\bm{x}|\bm{u})}{c(\bm{x})e(\bm{u})}$
 in~\eqref{univ-approx-exp-target} in spite of the contamination, which
 holds in the non-exponential family case as well (See
 Appendix~\ref{app:general-case} for more details). To this end, this
 section proposes two robust methods for nonlinear ICA based on the
 $\gamma$-cross entropy~\citep{fujisawa2008robust}, and shows that the
 desired estimation would be possible even under contamination by
 outliers.
 
 Before going to the details, let us clarify the notations:
 $p(\bm{x}|\bm{u})$ denotes the contaminated conditional density of
 $\bm{x}$ given $\bm{u}$ from~\eqref{source-contamination-model}, while
 $\pst(\bm{x}|\bm{u})$ and $\delta(\bm{x}|\bm{u})$ are the
 (noncontaminated) target and outlier conditional densities,
 respectively. We can obtain the two marginal densities from
 $p(\bm{x}|\bm{u})$ and $\pst(\bm{x}|\bm{u})$ as $p(\bm{x}):=\int
 p(\bm{x}|\bm{u}) p(\bm{u}) \intd\bm{u}$ and $\pst(\bm{x}):=\int
 \pst(\bm{x}|\bm{u})p(\bm{u})\intd\bm{u}$. In the remaining part of this
 paper, we may suppose that $p(\bm{u})$ is contaminated by an outlier
 density in general but the contaminated model is not explicitly defined
 because a specific form is not required in the analysis of this paper.
  \subsection{Nonlinear ICA  with robust binary classification}
  \label{ssec:binary}
  The first method performs nonlinear ICA by solving a binary
  classification problem under the $\gamma$-cross
  entropy~\citep{fujisawa2008robust,hung2018robust}. Let us express a
  class label by $y$, and $y=1$ and $y=0$ correspond to datasets $\calD$
  and $\calD_{\mathrm{p}}$ in~\eqref{binary-classification} which are
  drawn from $p(\bm{x},\bm{u}|y=1)=p(\bm{x},\bm{u})$ and
  $p(\bm{x},\bm{u}|y=0)=p(\bm{x})p(\bm{u})$, respectively.  Moreover,
  symmetric class probabilities are assumed (i.e.,
  $p(y=0)=p(y=1)=\frac{1}{2}$).  Then, the $\gamma$-cross entropy for
  binary classification is defined as
  \begin{align}
   d_{\gamma}(p(y|\bm{x},\bm{u}), r(\bm{x},\bm{u});
   p(\bm{x},\bm{u})) &:=-\frac{1}{\gamma}\log \iint
   \sum_{y=0}^{1}
   \left\{\frac{r(\bm{x},\bm{u})^{y(\gamma+1)}}{1+r(\bm{x},\bm{u})^{\gamma+1}}
   \right\}^{\frac{\gamma}{\gamma+1}} p(y,\bm{x},\bm{u})
   \intd\bm{x}\intd\bm{u}, \label{binary}
  \end{align}
  where $r(\bm{x},\bm{u})$ denotes a model (e.g., a neural network) and
  positive function.  As proven in~\citet{fujisawa2008robust}, the
  $\gamma$-cross entropy has a number of remarkable properties. For
  instance, $d_{\gamma}(p(y|\bm{x},\bm{u}), r(\bm{x},\bm{u});
  p(\bm{x},\bm{u}))$ approaches to the cross entropy in logistic
  regression as $\gamma\rightarrow0$. Notably, the $\gamma$-cross
  entropy has a desired robustness property on parameter estimation even
  in the heavy contamination of
  outliers~\citep{fujisawa2008robust,kanamori2015robust,kawashima2018difference}.
  Next, we show that the robust property holds in the context of
  nonlinear ICA.
  
  \paragraph{Robustness to heavy contamination of outliers in nonlinear ICA:} 
  First, we establish the following theorem to understand under what
  conditions a good estimation in the presence of outliers is possible
  for nonlinear ICA:
  \begin{theorem}
   \label{theo:robustness} Assume that
   \begin{align}
    \nu:=\iint \left\{\frac{r(\bm{x},\bm{u})^{\gamma+1}}
    {1+r(\bm{x},\bm{u})^{\gamma+1}}\right\}^{\frac{\gamma}{\gamma+1}}
    \epu\delta(\bm{x},\bm{u})\intd\bm{x}\intd\bm{u}
    \label{defi-nu}
   \end{align}
   is sufficiently small. Then, it holds that
   \begin{align}
    d_{\gamma}(p(y|\bm{x},\bm{u}), r(\bm{x},\bm{u});p(\bm{x},\bm{u}))
    &=J[r(\bm{x},\bm{u});(1-\epu)\pst(\bm{x},\bm{u}), p(\bm{x})p(\bm{u})]+O(\nu),
    \label{gamma-decomposition}
   \end{align}   
   where
   \begin{align*}
    &J[r(\bm{x},\bm{u});(1-\epu)\pst(\bm{x},\bm{u}), p(\bm{x})p(\bm{u})]\\
    &:=-\frac{1}{\gamma}\log \left[
    \frac{1}{2}\iint
    \left\{\frac{1}{1+r(\bm{x},\bm{u})^{\gamma+1}}\right\}^{\frac{\gamma}{\gamma+1}}
    p(\bm{x})p(\bm{u})\intd\bm{x}\intd\bm{u}
    +\frac{1}{2}\iint \left\{\frac{r(\bm{x},\bm{u})^{\gamma+1}}
    {1+r(\bm{x},\bm{u})^{\gamma+1}}\right\}^{\frac{\gamma}{\gamma+1}}
    (1-\epu)\pst(\bm{x},\bm{u})\intd\bm{x}\intd\bm{u}\right].
   \end{align*}
   Furthermore, under the assumption that $p(\bm{x})$, $p(\bm{u})$ and
   $r(\bm{x},\bm{u})$ are positive for all $\bm{x}$ and $\bm{u}$,
   $J[r(\bm{x},\bm{u});(1-\epu)\pst(\bm{x},\bm{u}),
   p(\bm{x})p(\bm{u})]$ is minimized at
   \begin{align}
    r^{\star}(\bm{x},\bm{u})
    &=\frac{(1-\epu)\pst(\bm{x}|\bm{u})}{p(\bm{x})}.
    \label{gamma-minimizer}
   \end{align}
  \end{theorem}    
  The proof is deferred to Appendix~\ref{app:robustness}.
  Theorem~\ref{theo:robustness} indicates that under the condition that
  $\nu$ is sufficiently small, we could obtain a desirable estimation
  result~\eqref{gamma-minimizer} in nonlinear ICA:
  Eq.\eqref{gamma-minimizer} is the special case of~the ideal universal
  approximation condition~\eqref{univ-approx-exp-target} without
  outliers where $c(\bm{x})=p(\bm{x})$ and $e(\bm{u})=1/(1-\epu)$. The
  notable point is that $\epu$ is never assumed to be small in
  itself. Thus, heavy contamination of outliers is also within the scope
  of our method.
    
  Let us define the supports of $\pst(\bm{s}|\bm{u})$ and
  $\delta(\bm{s}|\bm{u})$ as
  \begin{align*}
  \mathcal{S}^{\pst}_{\bm{u}}:=\{\bm{s}~|~\pst(\bm{s}|\bm{u})>0\},
   \quad
   \mathcal{S}^{\delta}_{\bm{u}}:=\{\bm{s}~|~\delta(\bm{s}|\bm{u})>0\},
  \end{align*}
  respectively.  Then, the following proposition gives insight into when
  $\nu$ is sufficiently small:
  \begin{proposition}
   \label{prop:nu} Let us denote the domains of $\bm{u}$ and $\bm{s}$ by
   $\mathcal{U}$ and $\mathcal{S}$, respectively. We assume that (i) the
   integrals in $\nu$ are defined over $\mathcal{U}$ and $\mathcal{S}$,
   (ii) data $\bm{x}$ is generated from~\eqref{ICA-model} with an
   invertible nonlinear mixing function $\bm{f}$, (iii)
   $\mathcal{S}^{\pst}_{\bm{u}}\cap\mathcal{S}^{\delta}_{\bm{u}}=\emptyset$,
   and (iv) $p(\bm{s})\neq0$ on $\mathcal{S}$. For $\gamma>0$,
   \begin{align*}
    \nu\leq 
    O\left(\sup_{\bm{x},\bm{u}}|r(\bm{x},\bm{u})-r^{\star}(\bm{x},\bm{u})|\right).
   \end{align*}
  \end{proposition}
  The proof is given in Appendix~\ref{app:proof-proposition}. The most
  important condition is
  $\mathcal{S}^{\pst}_{\bm{u}}\cap\mathcal{S}^{\delta}_{\bm{u}}=\emptyset$,
  which implies that $\pst(\bm{s}|\bm{u})$ and $\delta(\bm{s}|\bm{u})$
  are \emph{separated} on $\mathcal{S}$. For instance,
  $\pst(\bm{s}|\bm{u})$ and $\delta(\bm{s}|\bm{u})$ are the uniform
  densities on $[0,1]^{\dx}$ and $[2,3]^{\dx}$ respectively, their
  supports are nonoverlapping and thus separated. Therefore,
  Proposition~\ref{prop:nu} implies that $\nu$ can be sufficiently small
  in the neighborhood of $r^{\star}(\bm{x},\bm{u})$ when
  $\delta(\bm{s}|\bm{u})$ and $\pst(\bm{s}|\bm{u})$ are clearly
  \emph{separated}. This density separation would happen approximately
  on a situation where $\delta(\bm{s}|\bm{u})$ exists on the tails of
  $\pst(\bm{s}|\bm{u})$ as in a common outlier situation. 

  On the other hand, when $\gamma=0$ (i.e., logistic regression), it can
  be easily confirmed from the definition~\eqref{defi-nu} that $\nu$ is
  a nonzero constant and cannot be sufficiently small. Thus, nonlinear
  ICA methods based on logistic regression can be more sensitive to
  outliers. Appendix~\ref{app:neighbor-inequality} includes another
  discussion that the $\gamma$-entropy evades the influence of the large
  density ratio $\delta(\bm{s}|\bm{u})/\pst(\bm{s}|\bm{u})$, which is
  one of the main factors to hamper estimation in nonlinear ICA as
  already indicated in Section~\ref{sec:outliers}, while the logistic
  regression might not be able to avoid the strong influence from
  $\delta(\bm{s}|\bm{u})/\pst(\bm{s}|\bm{u})$.
  \paragraph{Influence function analysis:}
  Next, we investigate the robustness of our nonlinear ICA method based
  on the influence function (IF), which is an established measure in
  robust statistics~\citep{hampel2011robust}. To this end, let us define
  the following contaminated density model:
  \begin{align*}
   \bar{p}(\bm{x},\bm{u})
   &=(1-\epsilon)\pst(\bm{x},\bm{u})+\epsilon\bar{\delta}_{(\bx,\bu)}(\bm{x},\bm{u})\\
   \bar{p}(\bm{x})&=(1-\epsilon)\pst(\bm{x})+\epsilon\bar{\delta}_{\bx}(\bm{x})\\
   \bar{p}(\bm{u})&=(1-\epsilon)\pst(\bm{u})+\epsilon\bar{\delta}_{\bu}(\bm{u}),
  \end{align*}
  where $\epsilon\in[0,1)$ is a contamination ratio,
  $\bar{\delta}_{\bm{z}}$ is the Dirac delta function having a point
  mass at $\bm{z}$. We suppose that a model $r_{\vtheta}(\bm{x},\bm{u})$
  is positive and parameterized by $\vtheta$, and define $\vhtheta$ as a
  solution of the estimating function
  $\parder{\bm{\theta}}d_{\gamma}(\pst(y|\bm{x},\bm{u}),
  r_{\bm{\theta}}(\bm{x},\bm{u});\pst(\bm{x},\bm{u}))=\bm{0}$ over the
  (uncontaminated) target densities where
  $\pst(\bm{x},\bm{u}|y=1)=\pst(\bm{x},\bm{u})$ and
  $\pst(\bm{x},\bm{u}|y=0)=\pst(\bm{x})\pst(\bm{u})$. Similarly,
  $\vhtheta_{\epsilon}$ is defined as a solution of
  $\parder{\bm{\theta}}d_{\gamma}(\bar{p}(y|\bm{x},\bm{u}),
  r_{\bm{\theta}}(\bm{x},\bm{u});\bar{p}(\bm{x},\bm{u}))=\bm{0}$ over
  the contaminated densities where
  $\bar{p}(\bm{x},\bm{u}|y=1)=\bar{p}(\bm{x},\bm{u})$ and
  $\bar{p}(\bm{x},\bm{u}|y=0)=\bar{p}(\bm{x})\bar{p}(\bm{u})$. Then, IF
  is defined by
  \begin{align}
   {\rm IF}(\bx,\bu)=\lim_{\epsilon\to 0}
   \frac{\vhtheta-\vhtheta_{\epsilon}}{\epsilon}.
   \label{defi-IF}
  \end{align}
  Eq.\eqref{defi-IF} indicates that IF measures how $\vhtheta$ is
  influenced by outliers $(\bx,\bu)$ in the small contamination, and a
  larger IF implies that $\vhtheta$ is more sensitive to outliers.
  \emph{B-robustness} is a desired property for $\vhtheta$ in terms of
  IF: $\vhtheta$ is said to be B-robust when $\sup_{\bx,\bu}| {\rm
  IF}(\bx,\bu)|<\infty$~\citep{hampel2011robust}.  The following
  proposition shows that the solution of our method based on the
  $\gamma$-cross entropy can be B-robust under mild assumptions:
  \begin{proposition}
   \label{proposition.if1} Assume that a matrix $\bm{C}_{\bm{\theta}}$
   is invertible\footnote{The definition of $\bm{C}_{\bm{\theta}}$ is
   left in Appendix~\ref{app:proof-robust} because it is very
   complicated.}, and $r_{\vtheta}(\bm{x},\bm{u})$ satisfies
   \begin{align}
    \lim_{|r_{\vtheta}(\bm{x},\bm{u})|\to\infty}S_{\vtheta}(\bm{x},\bm{u})\frac{\partial
    \log\tf}{\partial \vtheta}=0, 
\label{condition:if1}
   \end{align}       
   where $S_{\vtheta}(\bm{x},\bm{u}):=\left\{
   L_{\vtheta}(\bm{x},\bm{u})(1-L_{\vtheta}(\bm{x},\bm{u}))\right\}^{\frac{\gamma}{1+\gamma}}$
   with $L_{\vtheta}(\bm{x},\bm{u}):=\frac{1}{1+\tf}$. Then, for
   $\gamma>0$,
   \begin{align}
    \sup_{\bx,\bu}|{\rm IF}(\bx,\bu)| <\infty.
   \end{align}
  \end{proposition}
  The proof is given in
  Appendix~\ref{app:proof-robust}. Proposition~\ref{proposition.if1}
  indicates that our method can be B-robust when
  $r_{\vtheta}(\bm{x},\bm{u})$ is modelled by a continuous yet unbounded
  function (possibly, feedforward neural networks with unbounded
  activation functions). Assumption~\eqref{condition:if1} is mild
  because when $|r_{\vtheta}(\bm{x},\bm{u})|$ diverges to infinite as
  $\|\bm{x}\|, \|\bm{u}\|\to\infty$ as in neural networks,
  $S_{\vtheta}(\bm{x},\bm{u})$ quickly approaches $0$.
  
  On the other hand, in the limit of $\gamma=0$ (i.e., logistic
  regression), a class of models for $r_{\vtheta}(\bm{x},\bm{u})$, which
  satisfies Assumption~\eqref{condition:if1}, is very limited because
  $S_{\vtheta}(\bm{x},\bm{u})=1$. For instance, when
  $r_{\vtheta}(\bm{x},\bm{u})$ is a neural network with an unbounded
  activation function, Assumption~\eqref{condition:if1} would not hold.
  Thus, our analysis implies that the $\gamma$-cross entropy is
  promising for nonlinear ICA in the presence of outliers.

    \paragraph{Robust permutation contrastive learning (RPCL):}
    As a practical method, we propose a robust variant of permutation
    contrastive learning (PCL)~\citep{pmlr-v54-hyvarinen17a} which we
    call \emph{robust permutation contrastive learning} (RPCL).  The
    original PCL supposes that sources are temporally dependent (e.g.,
    $\bm{s}(t)$ and $\bm{s}(t-1)$ are statistically dependent), and then
    makes use of the temporal dependencies for nonlinear ICA by
    regarding past information as the auxiliary variable
    $\bm{u}(t)=\bm{x}(t-1)$. PCL belongs to the non-exponential family
    case~\eqref{general-conditional-independence}.
    
    RPCL estimates a model $r(\bm{x},\bm{u})$ based on the following
    empirical $\gamma$-cross entropy for binary classification:
    \begin{align*}
     &\widehat{d}_{\gamma}(p(y|\bm{x},\bm{u}), r(\bm{x},\bm{u});p(\bm{x},\bm{u}))\\
     &:=-\frac{1}{\gamma}\log\left[ \frac{1}{2T}\sum_{t=1}^n
     \left\{\left(
     \frac{r(\bm{x}(t),\bm{u}(t))^{\gamma+1}}{1+r(\bm{x}(t),\bm{u}(t))^{\gamma+1}}
     \right)^{\frac{\gamma}{\gamma+1}}
     +\left(\frac{1}{1+r(\bm{x}(t),\bm{u}_{\mathrm{p}}(t))^{\gamma+1}}
     \right)^{\frac{\gamma}{\gamma+1}}\right\}\right], 
    \end{align*}
    where $\bm{u}_{\mathrm{p}}(t)$ denotes a random permutation of
    $\bm{u}(t)$ with respect to $t$. Based on the universal
    approximation assumption in~\citet[Theorem~1 and
    Eq.(12)]{pmlr-v54-hyvarinen17a} or Appendix~\ref{app:general-case},
    we restrict a model $r$ as
    $r(\bm{x}(t),\bm{u}(t))=\exp(\sum_{i=1}^{\dx}
    \psi_i(h_i(\bm{x}(t)),h_i(\bm{u}(t)))$ with a neural network
    $\bm{h}(\bm{x})=(h_1(\bm{x}),\dots,h_{\dx}(\bm{x}))^{\top}$.
    Following~\citet{pmlr-v54-hyvarinen17a},
    $\psi_i(h_i(\bm{x}),h_i(\bm{u}))$ was further modelled by
    $|a_{i,1}h_i(\bm{x})+a_{i,2}h_i(\bm{u})+b_i|
    -(\bar{a}_{i}h_i(\bm{x})+\bar{b}_i)^2+c$, where $a_{i,1}, a_{i,2},
    b_i, \bar{a}_i, \bar{b}_i, c$ are parameters to be estimated from
    data.  A minibatch stochastic gradient method is employed to
    optimize all parameters.
  \subsection{Nonlinear ICA  with robust multiclass classification}
  The second method is intended for the case where the auxiliary
  variable $u\in\{1,\dots,K\}$ is a one-dimensional and $K$-discrete
  variable (e.g., class label).  To this end, we solve a multiclass
  classification problem based on the $\gamma$-cross entropy:
  \begin{align}
   &d_{\gamma}(p(u|\bm{x}),r(u,\bm{x});p(\bm{x}))
   :=-\frac{1}{\gamma}\log \int\left\{ \frac{\sum_{u=1}^K
   r(u,\bm{x})^{\gamma}p(u|\bm{x})} {\left(\sum_{u'=1}^K
   r(u',\bm{x})^{\gamma+1}\right)^{\frac{\gamma}{\gamma+1}}}\right\}
   p(\bm{x})\intd\bm{x}, \label{multiclass}
  \end{align}    
  where we supposed $p(u=1)=p(u=2)=\dots=p(u=K)$.  Regarding multiclass
  classification, a robustness property similar to what we had above
  holds by modifying the above discussion on binary
  classification~\eqref{binary} or
  following~\citet{kawashima2018difference}. When $\pst(\bm{s}|u)$ and
  $\delta(\bm{s}|u)$ are clearly \emph{separated}, minimization of
  $d_{\gamma}(p(u|\bm{x}),r(u,\bm{x});p(\bm{x}))$ would enable us to
  estimate $\pst(\bm{x}|u)$, which is an ideal estimation result and a
  special case of $\frac{\pst(\bm{x}|u)}{c(\bm{x})e(u)}$
  in~\eqref{univ-approx-exp-target}.  Details are given in
  Appendix~\ref{app:multi-robustness}.
  
    \paragraph{Robust time contrastive learning (RTCL):}    
    As a practical method in multiclass classification, we propose
    \emph{robust time contrastive learning} (RTCL) which is a robust
    version of TCL based on the $\gamma$-cross
    entropy~\eqref{multiclass}. Both TCL and RTCL are intended for the
    conditional independent exponential family
    case~\eqref{exp-family-TCL}, and suppose time series data
    (artificially or manually) divided into $K$ time segments, and the
    auxiliary variable $u\in\{1,\dots,K\}$ is the time segment label.
    RTCL employs the following empirical $\gamma$-cross entropy:
    \begin{align*}
     &\widehat{d}_{\gamma}(p(u|\bm{x},\bm{u}), 
     r(\bm{x},u);p(\bm{x}))
     :=-\frac{1}{\gamma}\log \left[\frac{1}{T}\sum_{t=1}^T
     \left\{\frac{\left(\sum_{k=1}^K
     \delta_{u(t),k}~r(u(t),\bm{x}(t))^{\gamma}\right)}
     {\left(\sum_{u'=1}^Kr(u',\bm{x}(t))^{\gamma+1}\right)^{\frac{\gamma}{\gamma+1}}}
     \right\}\right],
    \end{align*}
    where $u(t)\in\{1,\dots,K\}$ are the observations of time-segment
    labels, and $\delta_{u(t),k}$ denotes the Kronecker delta. Based on
    the universal approximation assumption (A4) in
    Theorem~\ref{theo:identifiability}, we restrict $r$ as
    $r(\bm{x},u)=\exp(\bm{w}_u^{\top}\bm{h}(\bm{x})+b_u)$ for
    $u=1,\dots,K$ where $\bm{h}(\bm{x})$ denotes nonlinear ICA features
    modelled by a neural network, and $\bm{w}_u$ and $b_u$ are
    parameters for weights and bias, respectively. In practice, all
    parameters are optimized by a minibatch stochastic gradient method.
 \section{Numerical experiments on artificial data}
 \label{sec:artdata}
 This section numerically investigates the robustness of RTCL and RPCL
 with comparison to existing nonlinear ICA methods on artificial data.
  \subsection{Robust time contrastive learning}
  \label{ssec:artdata_TCL}
  \begin{figure}
   \centering
   \subfigure[$q(s)$]{\includegraphics[width=.9\textwidth]
   {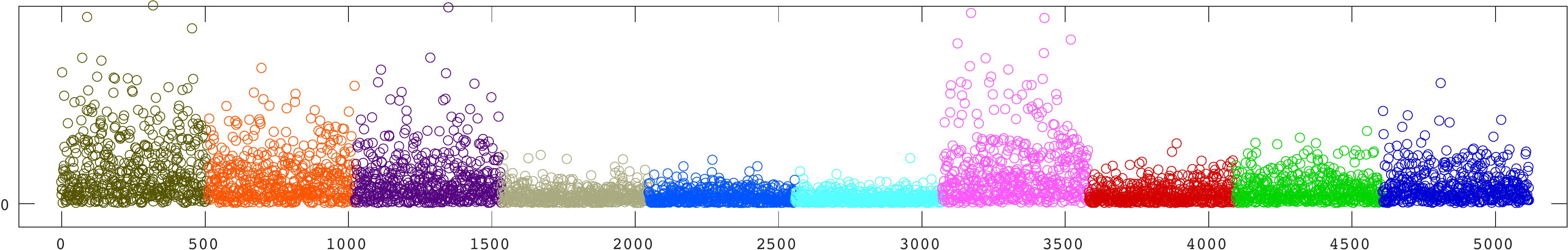}} 
   \subfigure[RCTL
   ($\gamma=1.0$)]{\includegraphics[width=.9\textwidth]{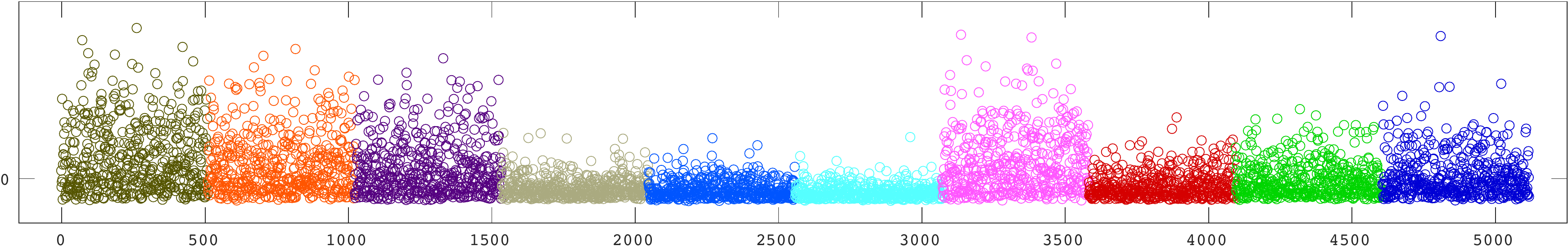}}
   \subfigure[TCL]{\includegraphics[width=.9\textwidth]{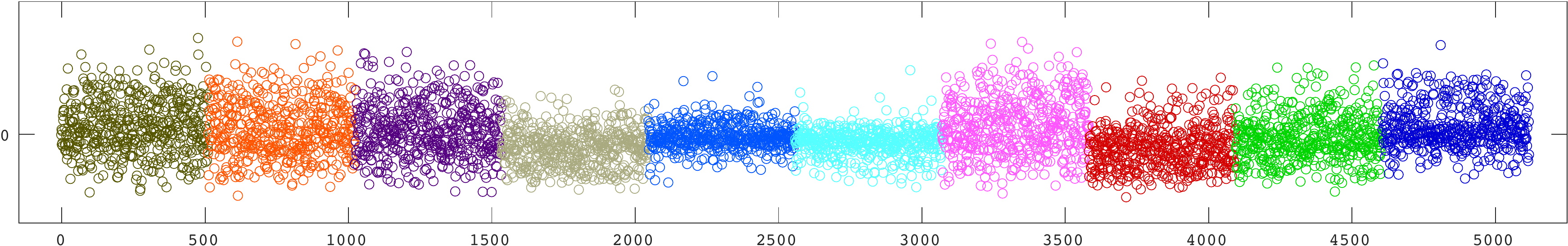}}
   \caption{\label{fig:demo} (a) $q(s)=|s|$ without outliers and ICA
   features $h(\bm{x})$ estimated by (b) RTCL and (c) TCL on ten time
   segments.  Each time segment is highlighted by a color. The outlier
   density is the Laplace density, and $\epsilon=0.1$.}
  \end{figure}
  \begin{table*}[t]  
  \begin{center}
   \caption{\label{exp-TCL1} TCL and TCL on artificial data.  Averages
   of mean absolute correlations are computed over $10$ runs. The
   outlier densities are the independent Laplace density and the
   modulated mixture of two Gaussians in the top and bottom panels,
   respectively.  A larger value indicates a better result. The best and
   comparable methods judged by the t-test at the significance level 5\%
   are described in boldface.}
   \vspace{2mm}
   \begin{tabular}{l|c|c|c|c|c}
    \hline Laplace  & TCL & RTCL ($\gamma=0.1$) & RTCL ($\gamma=0.3$) & RTCL ($\gamma=0.5$) & RTCL ($\gamma=1$) \\ \hline 
    $\epsilon=0.01$ & 0.890(0.010) & 0.933(0.011) & 0.974(0.009) & {\bf 0.985(0.010)} & {\bf 0.993(0.010)}\\
    $\epsilon=0.03$ & 0.810(0.017) & 0.865(0.013) & 0.919(0.015) & 0.947(0.015) & {\bf 0.976(0.014)}\\
    $\epsilon=0.05$ & 0.751(0.039) & 0.800(0.026) & 0.869(0.012) & 0.898(0.017) & {\bf 0.947(0.019)}\\
    $\epsilon=0.1$ & 0.627(0.032) & 0.690(0.058) & 0.787(0.023) & 0.832(0.024) & {\bf 0.893(0.014)}
   \end{tabular}
  \end{center}
  \begin{center}
   \vspace{1mm}   
   \begin{tabular}{l|c|c|c|c|c}
    \hline Gaussian &  TCL & RTCL ($\gamma=0.1$) & RTCL ($\gamma=0.3$) & RTCL
   ($\gamma=0.5$) & RTCL ($\gamma=1$) \\ \hline $\epsilon=0.01$ &
   0.948(0.005) & 0.981(0.007) & 0.992(0.004) & 0.995(0.002) & {\bf
   0.997(0.002)}\\ $\epsilon=0.03$ & 0.751(0.030) & 0.876(0.033) &
   0.949(0.009) & 0.968(0.007) & {\bf 0.980(0.007)}\\ $\epsilon=0.05$ &
   0.724(0.036) & 0.753(0.046) & 0.876(0.041) & 0.924(0.022) & {\bf
   0.958(0.011)}\\ $\epsilon=0.1$ & 0.778(0.035) & 0.757(0.035) &
   0.822(0.026) & 0.862(0.012) & {\bf 0.914(0.012)}
   \end{tabular}
  \end{center}
  \end{table*}
    \paragraph{Data generation, nonlinear ICA methods, evaluation:} 
    We slightly modified the experimental setting of
    TCL\footnote{\url{https://github.com/hirosm/TCL}}
    in~\citet{hyvarinen2016unsupervised}. Let us note that TCL is a
    nonlinear ICA method specialized in the conditionally independent
    exponential family~\eqref{exp-family-TCL}. Source vectors with time
    segment length $512$ was first generated
    from~\eqref{source-contamination-model}:
    Following~\eqref{exp-family-TCL}, given a time segment label,
    $\pst(\bm{s}|u)$ was conditionally independent Laplace distributions
    with means $0$ and different scales across time segments, which were
    randomly determined from the uniform distribution on
    $[0,\frac{1}{\sqrt{2}}]$. Regarding the outlier density
    $\delta(\bm{s}|u)$, two types of densities were used: An independent
    Laplace distribution with mean $0$ and scale $3.0$, and a modulated
    mixture of two Gaussians where the mixing coefficients are $0.5$ and
    the standard deviations are fixed at $0.5$, but the mean parameters
    depend on the time segments $u$ and were randomly determined by the
    uniform density on $[1.0, 4.0]$ and $[-4.0, -1.0]$, respectively.
    We set $\epsilon(u)=\epsilon$ for all time segments $u$. The total
    numbers of segments and of data samples were $K=256$ and
    $T=512\times256$ respectively, while the dimensionality of data was
    $\dx=10$. Then, data $\bm{x}$ was generated according
    to~\eqref{ICA-model} where $\bm{f}(\bm{s})$ was modelled by a
    three-layer neural network with the leaky ReLU activation function
    and random weights. The numbers of all hidden and output units were
    the same as the dimensionality of data. As preprocessing, we
    performed whitening based on the
    $\gamma$-divergence~\citep{chen2013robust}.

    ICA features $\bm{h}(\bm{x})$ both in RTCL and TCL were modelled by
    a three layer neural network where the number of hidden units was
    $4\dx$, but the final layer was $\dx$. Regarding the activation
    functions, the final layer employed the absolute value function,
    while the other hidden layers were the max-out
    function~\citep{goodfellow2013maxout} with two groups. $\ell_2$
    regularization was employed with the regularization parameter
    $10^{-4}$.  When $\epsilon<0.1$, we optimized the network parameters
    both in RTCL and TCL with $0.001$ learning rate using the Adam
    optimizer~\citep{kingma2015adam} for $1,000$ epochs with mini-batch
    size $256$, while we updated the parameters for $3,000$ epochs for
    $\epsilon=0.1$. Regarding RTCL, to avoid bad local optima, we
    initialized the network parameters in RTCL by the parameters
    optimized by TCL for $300$ epochs.  As postprocessing,
    FastICA~\citep{hyvarinen1999fast} was finally applied to the
    estimated features $\bm{h}(\bm{x})$ both in TCL and RTCL as done
    in~\citet{hyvarinen2016unsupervised}.
    
    When evaluating the performances of TCL and RTCL, sources without
    outliers were used. Then, the performance was measured by the
    Pearson correlation coefficient between the the absolute values of
    the estimated sources and of the original sources (i.e.,
    $|\bm{h}(\bm{x})|$ after FastICA and $\bm{q}(\bm{s})$).
    \paragraph{Results:}       
    Fig.\ref{fig:demo} illustrates examples of the ICA features
    $\bm{h(\bm{x})}$ by RTCL and TCL. Compared with the original
    $q(s)=|s|$ (one-dimensional source) (Fig.\ref{fig:demo}(a)), RTCL
    captures the nonstationary of the original source and well-recovers
    $q(s)$ (Fig.\ref{fig:demo}(b)). On the other hand, TCL does not
    recover $q(s)$ at all (Fig.\ref{fig:demo}(c)). This result supports
    the conclusion of Theorem~\ref{theo:identifiability} that TCL can be
    sensitive to outliers.    

    The top panel in Table~\ref{exp-TCL1} quantitatively indicates that
    RTCL is more robust against outliers than TCL. As the contamination
    ratio $\epsilon$ increases, the performance of TCL deteriorates.
    On the other hand, RTCL keeps high-correlation values even for
    larger $\gamma$. When the outlier density $\delta(\bm{x}|u)$ is the
    modulated mixture of two Gaussians, RTCL still performs well (bottom
    panel in Table~\ref{exp-TCL1}).
  \subsection{Robust permutation contrastive learning}
  \label{ssec:artdata_PCL}
    \paragraph{Data generation, nonlinear ICA methods, evaluation:}     
    We followed the experimental setting of PCL
    in~\citet{pmlr-v54-hyvarinen17a}. Let us note that PCL is a special
    case of nonlinear ICA for the non-exponential
    case~\eqref{general-conditional-independence}. First, the temporally
    dependent $T$ sources were generated from
    \begin{align*}
     \log\pst(\bm{s}(t)|\bm{s}(t-1))=-\sum_{i=1}^{\dx} |s_i(t)-\rho
     s_i(t-1)|+C
    \end{align*}
    where $C$ denotes a constant and the auto-regressive coefficient
    $\rho$ was fixed at $0.7$. The total number of sources was
    $T=65536$. Then, we randomly replaced the sources by outliers based
    on a constant contamination ratio $\epsilon$, which were generated
    from the independent Laplace density with the mean $0$ and scale
    $3$. To generate $\bm{x}$, the sources with outliers are nonlinearly
    mixed by the same neural networks as previous experiments.
    
    ICA features $\bm{h}(\bm{x})$ both in RPCL and PCL were modelled by
    the same neural networks as in the experiments for RTCL except for
    that no activation function was applied to the outputs on the last
    layer.  $\ell_2$ regularization was employed with the regularization
    parameter $10^{-4}$.  Then, we optimized the parameters in RPCL and
    PCL with $0.001$ learning rate using Adam for $1,000$ epochs with
    mini-batch size $128$.\footnote{Regarding only for a few runs on the
    whole experiments, the learning rate in RPCL was decreased as the
    number of iterations increased for numerical stability} Regarding
    RPCL, to avoid bad local optima, all parameters in RPCL are
    initialized by the parameters optimized by PCL for $100$ epochs.
    Unlike the experiments in RTCL, no postprocessing was applied.  The
    performance was measured by the Pearson correlation coefficient
    between learned ICA features $\bm{h}(\bm{x})$ and $\bm{s}$ without
    outliers.
    \paragraph{Results:}       
    Table~\ref{exp-PCL} clearly shows that the correlation for PCL
    quickly decreases as the contaminating ratio $\epsilon$ increases.
    On the other hand, RPCL works significantly better than PCL even for
    large $\epsilon$. Thus, our methods based on the $\gamma$-cross
    entropy are promising.
    \begin{table*}[t]
     \begin{center}
      \caption{\label{exp-PCL} RPCL and PCL on artificial data. Averages
      of mean absolute correlations are computed over $10$ runs.}
      \vspace{2mm}
      \begin{tabular}{l|c|c|c|l|c}
       \hline    & PCL & RPCL ($\gamma=0.5$) & RPCL ($\gamma=1$) & RPCL ($\gamma=5$) & RPCL ($\gamma=10$) \\ \hline 
       $\epsilon=0.01$ & 0.917(0.028) & {\bf 0.935(0.010)} & {\bf 0.942(0.023)} & {\bf 0.934(0.026)} & 0.911(0.027)\\
       $\epsilon=0.05$ & 0.904(0.022) & {\bf 0.917(0.015)} & {\bf 0.926(0.008)} & {\bf 0.932(0.024)} & 0.899(0.030)\\
       $\epsilon=0.1$ & 0.854(0.053) & 0.884(0.034) & {\bf 0.888(0.029)} & {\bf 0.912(0.026)} & 0.886(0.023)\\
       $\epsilon=0.15$ & 0.803(0.058) & 0.819(0.056) & {\bf 0.838(0.048)} & {\bf 0.851(0.056)} & {\bf 0.866(0.031)}
      \end{tabular}
     \end{center}   
    \end{table*}

 \section{Application to causal discovery of Hippocampal fMRI data}
 An important application of nonlinear ICA is causal discovery whose
 objective is to learn the causal structure from observed data without
 relying on interventions~\citep{pearl2000causality}. The use of linear
 ICA methods is well-established in causal
 discovery~\citep{shimizu2006linear}, and TCL has also been employed
 recently~\citep{monti2019causal}.
 
\begin{figure}[t]
  \centering \includegraphics[width=.55\textwidth]{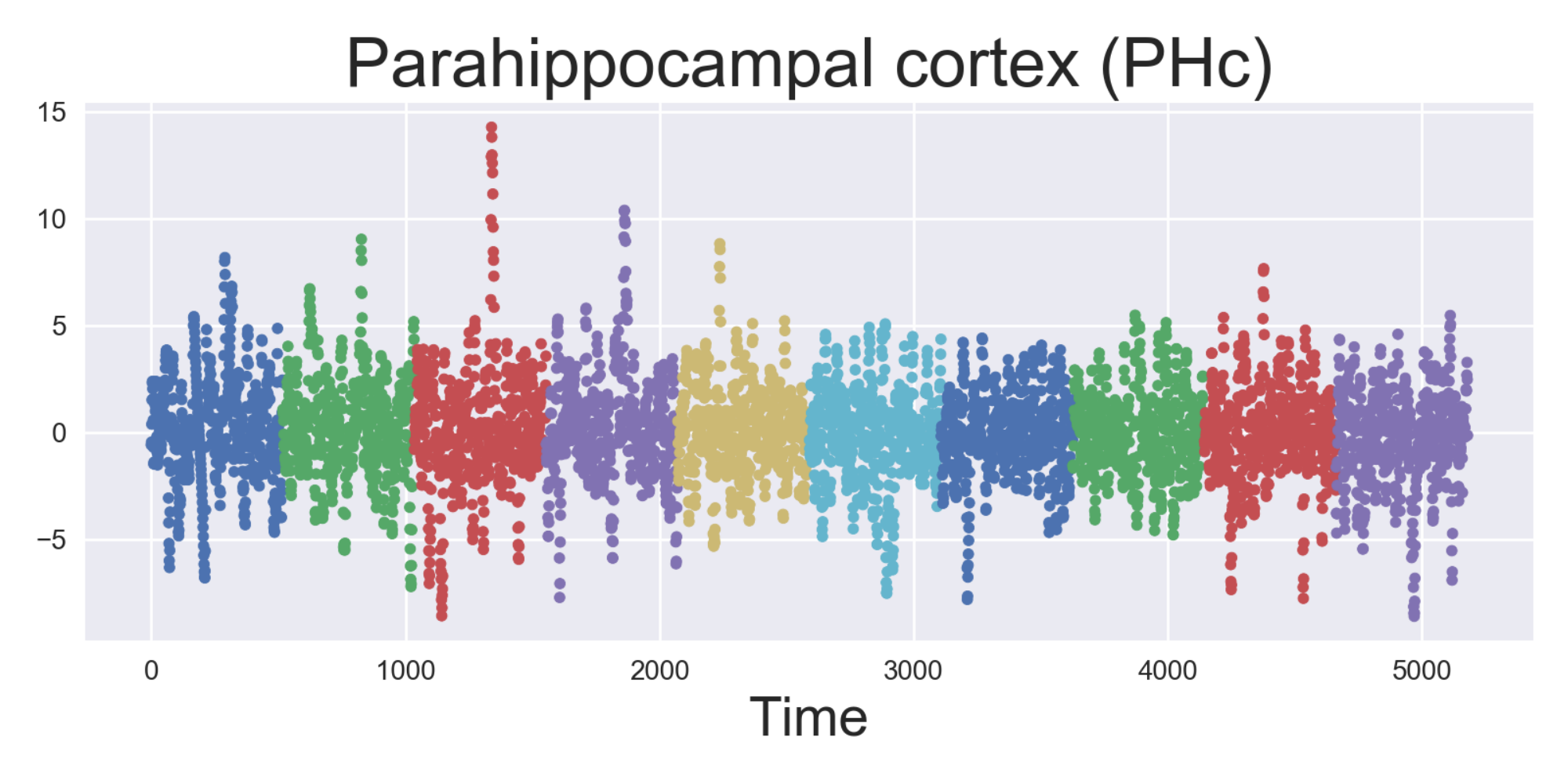}
  \caption{\label{PHcFigure} Subset of time series data corresponding to
  the Parahippocampal (PHc) region taken from the Hippocampal fMRI
  dataset.  Different colors denote distinct segments, which in this
  case correspond to fMRI measurements from the same subject on distinct
  days.}
 \end{figure}

 To demonstrate its applicability on a realworld dataset, we apply RTCL
 to causal discovery on resting-state fMRI data, which is well-known to
 contain outliers due to measurement issues such as head movement and
 variability in vascular health across a cohort of
 subjects~\citep{poldrack2011handbook}.  The dataset we consider
 corresponds to resting state fMRI data collected from a single subject
 (caucasian male, 45 years old) over 84 successive
 days~\citep{poldrack2015long}. Here, each day is treated as a distinct
 experimental condition. Fig.\ref{PHcFigure} visualizes the presence of
 outliers in the time series data for the Parahippocampal brain region.

 We follow a TCL-based method for nonlinear causal
 discovery~\citep{monti2019causal}. Let us consider the problem of
 causal discovery for bivariate data $\bm{x}=(x_1,x_2)^{\top}$. The goal
 of causal discovery is to determine whether $x_1$ causes $x_2$ or $x_2$
 causes $x_1$ (i.e., $x_1 \rightarrow x_2$ or $x_2 \rightarrow x_1$), or
 to conclude that no acyclic causal relation exists.  If the true causal
 direction is $x_1 \rightarrow x_2$, the (possibly) structural equation
 model (SEM)~\citep{pearl2000causality} can be written as
 \begin{align}
  x_1 &= f_1(n_1), \qquad
  x_2 = f_2(x_1, n_2), \label{SEM}
 \end{align}
 where $n_1$ and $n_2$ are latent disturbances and assumed to be
 statistically independent each other. As discussed
 in~\citet{monti2019causal}, the nonlinear SEM~\eqref{SEM} has a clear
 connection to the data generative model~\eqref{ICA-model} in nonlinear
 ICA. Roughly, the disturbance variables $(n_1, n_2)^{\top}$ in SEM
 corresponds the latent sources $(s_1, s_2)^{\top}$ in ICA up to their
 permutation. Thus, regarding the recovered sources by nonlinear ICA as
 estimates of $(n_1, n_2)^{\top}$, we could determine the causal
 direction by performing a series of independence tests with the
 observations of $\bm{x}=(x_1, x_2)^{\top}$. For instance, under the
 assumption that the true causal direction is $x_1 \rightarrow x_2$, we
 need to verify that $x_1 \bigCI n_2$ while $x_1 \nbigCI n_1$, $x_2
 \nbigCI n_1$ and $x_2 \nbigCI n_2$~\citep[Property~1]{monti2019causal}
 by applying some independent test where $\bigCI$ (or $\nbigCI$) denotes
 statistical independence (or dependence). Here, we employed
 Hilbert-Schmidt independence criteria~\citep{gretton2005measuring} for
 independence test. This approach for bivariate data is extended to
 multivariate data in~\citet[Section~3.5]{monti2019causal}.
 
\begin{figure}[t]
   \centering
  \includegraphics[width=.9\textwidth]{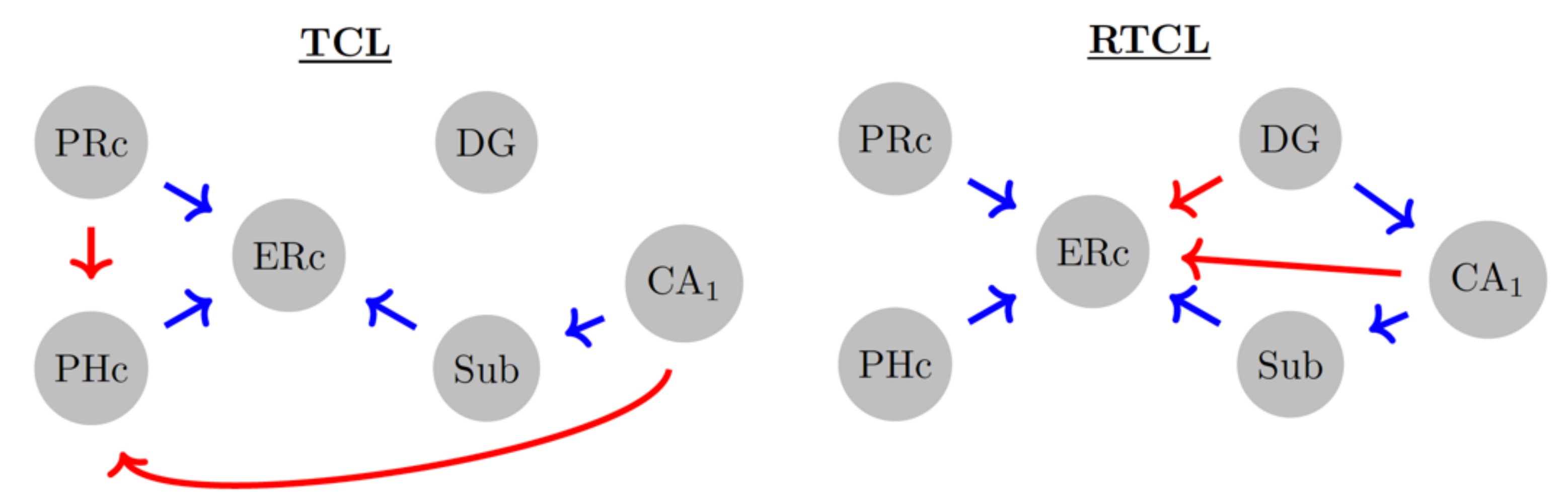}
  \caption{\label{hippocampusDAGs} Estimated directed acyclic graphs
  based on TCL (left panel) and RTCL (right panel, $\gamma=2.5$). For
  RTCL, the $\gamma$ value was selected based on classification accuracy
  for validation data.}
  \end{figure}

 Fig.~\ref{hippocampusDAGs} shows the causal structures obtained via TCL
 by~\citet{monti2019causal}, and RTCL where a five layer neural network
 is used.  Blue arrows denote edges which are plausible given the
 anatomical connectivity, while red arrows are not compatible with the
 known anatomical structure. We note that in the case of RTCL, the
 erroneous edges (highlighted in red) actually correspond to indirect
 causal effects. For example, see the edge between the Cornu Ammonis 1
 (CA$_1$) node and the entorhinal cortex (ERc) node. While a direct
 connection between such nodes is anatomically implausible, there is an
 indirect effect which is mediated by the subiculum (Sub) node.  This is
 in stark contrast with the results provided by TCL, where erroneous
 edges (highlighted in red) are not compatible with the anatomical
 structure (e.g., the TCL edge between CA$_1$ and PHc cannot be
 explained as an indirect causal effect).
 \section{Conclusion}
 We first analyzed the estimation of nonlinear ICA models in the
 presence of outliers, and then proposed two robust methods for
 nonlinear ICA. We showed by theoretical analysis that our methods have
 robustness properties in the context of nonlinear ICA. The robustness
 was further empirically shown in simulations, and applicability to
 real-data was also demonstrated through causal discovery.
 \subsection*{Acknowledgement}
 The authors would like to thank Dr. Hiroshi Morioka for sharing his PCL
 codes with us.
 \appendix
 \section{Proof of Theorem~\ref{theo:identifiability}}
 \label{app:identifiability}
 \begin{proof}
  Let us express the inverse of $\bm{f}$ in
  \eqref{source-contamination-model} by $\bm{g}$ such that
  $\bm{s}=\bm{g}(\bm{x})$. The change of variables provides
  \begin{align*}
   \log p(\bm{x}|\bm{u})&=\log\left\{(1-\epu) \pst(\bm{g}(\bm{x})|\bm{u})
   +\epu\delta(\bm{g}(\bm{x})|\bm{u})\right\}+\log |\det\bm{J_g}(\bm{x})|\\
   &=\log\left\{\pst(\bm{g}(\bm{x})|\bm{u})
   +\epu(\delta(\bm{g}(\bm{x})|\bm{u})-\pst(\bm{g}(\bm{x})|\bm{u}))\right\}
   +\log |\det\bm{J_g}(\bm{x})|\\
   &=\log\left[\pst(\bm{g}(\bm{x})|\bm{u})\left\{
   1+\epu\left(\frac{\delta(\bm{g}(\bm{x})|\bm{u})}{\pst(\bm{g}(\bm{x})|\bm{u})}-1\right)\right\}\right]
   +\log |\det\bm{J_g}(\bm{x})|\\
   &=\log \pst(\bm{g}(\bm{x})|\bm{u})
   +\epu\left(\frac{\delta(\bm{g}(\bm{x})|\bm{u})}
   {\pst(\bm{g}(\bm{x})|\bm{u})}-1\right)
   +O(\epu^2)+\log |\det\bm{J_g}(\bm{x})|,
  \end{align*}
  where we applied $\log(1+\epu z)=\epu z+O(\epu^2)$ with a sufficiently
  small $\epu$ on the last line. Then, the conditionally exponential
  family assumption~(A2) gives
  \begin{align*}
   \log p(\bm{x}|\bm{u})&=\sum_{j=1}^d \lambda_{j}(\bm{u})q_j(g_j(\bm{x}))
   +\lambda_{0}(\bm{u})+\epu\left(\frac{\delta(\bm{g}(\bm{x})|\bm{u})}
   {\pst(\bm{g}(\bm{x})|\bm{u})}-1\right)
   +O(\epu^2)+\log |\det\bm{J_g}(\bm{x})|-\log Z(\bm{\lambda}(\bm{u})).
  \end{align*}

  Contrasting two log-conditional densities of $\bm{x}$ given $\bm{u}$
  and an arbitrary fixed point $\bm{u}_0$ yields
  \begin{align}
   \log{p}(\bm{x}|\bm{u})-\log{p}(\bm{x}|\bm{u}_0)
   &=(\bm{\lambda}(\bm{u})-\bm{\lambda}(\bm{u}_0))^{\top}\bm{q}(\bm{g}(\bm{x}))
   +\left\{\epu\frac{\delta(\bm{g}(\bm{x})|\bm{u})}{\pst(\bm{g}(\bm{x})|\bm{u})}
   -\epsilon(\bm{u}_0)
   \frac{\delta(\bm{g}(\bm{x})|\bm{u}_0)}{\pst(\bm{g}(\bm{x})|\bm{u}_0)}\right\}
   \nonumber\\ &+O(\epu^2)+O(\epsilon(\bm{u}_0)^2) -(\log
   Z(\bm{\lambda}(\bm{u}))-\log Z(\bm{\lambda}(\bm{u}_0))),
   \label{contrast-pdfs}
  \end{align}
  where
  $\bm{\lambda}(\bm{u}):=(\lambda_{0}(\bm{u}),\dots,\lambda_{\dx}(\bm{u}))^{\top}$,
  $\bm{q}(\bm{g}(\bm{x})):=(q_{1}(g_1(\bm{x})),\dots,q_{\dx}(g_{\dx}(\bm{x})))^{\top}$,
  and note that the Jacobian $|\det\bm{J_g}(\bm{x})|$ is cancelled out.
  
  On the other hand, by the universal approximation assumption (A4), we
  obtain
  \begin{align}
   \log{p}(\bm{x}|\bm{u})-\log{p}(\bm{x}|\bm{u}_0)
   =(\bm{w}(\bm{u})-\bm{w}(\bm{u}_0))^{\top}\bm{h}(\bm{x})-e(\bm{u})+e(\bm{u}_0).
   \label{contrast-NN}
  \end{align}
  Then, equating~\eqref{contrast-pdfs} with \eqref{contrast-NN} provides
  \begin{align}
   \bar{\bm{\lambda}}(\bm{u})^{\top}\bm{q}(\bm{g}(\bm{x}))
   +\epu\frac{\delta(\bm{g}(\bm{x})|\bm{u})}{\pst(\bm{g}(\bm{x})|\bm{u})}
   -\epsilon(\bm{u}_0)\frac{\delta(\bm{g}(\bm{x})|\bm{u}_0)}{\pst(\bm{g}(\bm{x})|\bm{u}_0)}
   +O(\epu^2)+O(\epsilon(\bm{u}_0)^2)
   =\bar{\bm{w}}(\bm{u})^{\top}\bm{h}(\bm{x})+\bar{\beta}(\bm{u}),
   \label{eq-contrast}
  \end{align}
  where $\bar{\bm{w}}(\bm{u}):=\bm{w}(\bm{u})-\bm{w}(\bm{u}_0)$,
  $\bar{\bm{\lambda}}(\bm{u}):=\bm{\lambda}(\bm{u})-\bm{\lambda}(\bm{u}_0)$,
  and $\bar{\beta}(\bm{u}):= e(\bm{u})-e(\bm{u}_0)+\log
  Z(\bm{\lambda}(\bm{u})) -\log Z(\bm{\lambda}(\bm{u}_0))$.

  Next, we multiply $\bar{\bm{\lambda}}(\bm{u})$ to the both sides
  of~\eqref{eq-contrast} and evaluate it at $m$ points,
  $\bm{u}_1,\dots,\bm{u}_m$. Finally, taking the summation for
  $\bm{u}_1,\dots,\bm{u}_m$ yields
  \begin{align*}
   &\underbrace{\left(\sum_{i=1}^m
   \bar{\bm{\lambda}}(\bm{u}_i)
   \bar{\bm{\lambda}}(\bm{u}_i)^{\top}
   \right)}_{\bar{\bm{\Lambda}}}\bm{q}(\bm{s})
   +\sum_{i=1}^m
   \left(\epsilon(\bm{u}_i)
   \frac{\delta(\bm{g}(\bm{x})|\bm{u}_i)}{\pst(\bm{g}(\bm{x})|\bm{u}_i)}
   -\epsilon(\bm{u}_0)
   \frac{\delta(\bm{g}(\bm{x})|\bm{u}_0)}{\pst(\bm{g}(\bm{x})|\bm{u}_0)}\right)
   \bar{\bm{\lambda}}(\bm{u}_i)\\
   &\qquad+\sum_{i=1}^m \left\{O(\epsilon(\bm{u}_i)^2)+\epsilon(\bm{u}_0)^2)\right\}\bar{\bm{\lambda}}(\bm{u}_i)   
   =\left(\sum_{i=1}^m\bar{\bm{w}}(\bm{u}_i)\bar{\bm{\lambda}}(\bm{u}_i)^{\top}
   \right)\bm{h}(\bm{x})
   +\sum_{i=1}^m\bar{\beta}(\bm{u}_i)\bar{\bm{\lambda}}(\bm{u}_i).
  \end{align*}
  Applying the inverse of $\bar{\bm{\Lambda}}$ to both sides completes
  the proof.
  \end{proof}
 \section{Influence of outliers in general non-exponential case} 
 \label{app:general-case}
 As in~\citet{hyvarinen2018nonlinear}, we assume that the general
 conditional independence~\eqref{general-conditional-independence}
 holds, and that both the mixing function $\bm{f}$ in the data
 generative model~\eqref{ICA-model} and a nonlinear feature
 $\bm{h}(\bm{x})=(h_1(\bm{x}),\dots,h_{\dx}(\bm{x}))^{\top}$ are
 invertible. Then, by the change of variables from the generative
 model~\eqref{ICA-model}, $\bm{s}$ can be regarded as
 $\bm{h}:=\bm{h}(\bm{x})$:
  \begin{align*}
   \bm{s}(\bm{h}):=\bm{s}
   =\bm{f}^{-1}(\bm{x})=\bm{f}^{-1}\circ\bm{h}^{-1}(\bm{h}(\bm{x})),
  \end{align*}
  where $\circ$ denotes composition. For the case of no outliers,
  \citet{hyvarinen2018nonlinear} proved that the nonlinear feature
  $\bm{h}(\bm{x})$ asymptotically recover the source $\bm{s}$ up to an
  invertible transformation by confirming that the following conditions
  hold: For all $i,j,k=1,\dots,\dx$ but $j\neq k$,
  \begin{align}
   \parder{h_j}s_{i}(\bm{h})\parder{h_k}s_{i}(\bm{h})=0,~
   \text{and}~
   \frac{\partial^2}{\partial h_j\partial h_k}s_{i}(\bm{h})=0.
   \label{general-condition}
  \end{align}
  Eq.\eqref{general-condition} indicates that each $s_i$ is a function
  of only one distinct element in the nonlinear feature vector $\bm{h}$,
  and thus $\bm{s}$ is identifiable by $\bm{h}(\bm{x})$ up to an
  invertible transformation. Conditions~\eqref{general-condition} for
  all $i,j,k$ can be compactly expressed as the following matrix form:
  \begin{align}
   \bm{M}(\bm{s}(\bm{h}))=\bm{O}, \label{general-condition-matrix}
  \end{align}
  where $\bm{O}$ is the null matrix, and $\bm{M}(\bm{s}(\bm{h}))$ is a
  $(\dx^2-\dx)$ by $2\dx$ matrix with elements
  $\parder{h_j}s_{i}(\bm{h})\parder{h_k}s_{i}(\bm{h})$ and
  $\frac{\partial^2}{\partial h_j\partial h_k}s_{i}(\bm{h})$ for all
  $i,j,k=1,\dots,\dx$ but $j\neq k$ ($i$ for columns, and $j,k$ for rows
  in $\bm{M}(\bm{s}(\bm{h}))$).

  However, in the presence of outliers, \eqref{general-condition-matrix}
  does not hold in general. To investigate the influence from outliers,
  let us define $\widetilde{\bm{M}}(\bm{s}(\bm{h}))$ by a $(\dx^2-\dx)$
  by $(\dx^2-\dx)$ matrix whose elements are given by
  \begin{align*}
   -\parder{h_j}s_{l}(\bm{h})\parder{h_k}s_{m}(\bm{h}),
   ~\text{for all
   $j,k,l,m=1,\dots,\dx$ but $j\neq k$ and $l\neq m$}.
  \end{align*}
  Unlike $\bm{M}(\bm{s}(\bm{h}))$, $\widetilde{\bm{M}}(\bm{s}(\bm{h}))$
  includes the cross terms related to $s_l$ and $s_m$ for $l\neq m$. The
  following theorem is useful to understand how
  \eqref{general-condition-matrix} is modified by outliers:
  \begin{theorem}   
   \label{theo-general-case} Assume that
   \begin{enumerate}[({B}1)]
    \item Data $\bm{x}$ is generated from~\eqref{ICA-model} where
  	  $\bm{f}$ is invertible.
	  
    \item Conditional
  	  independence~\eqref{general-conditional-independence} holds in
  	  $\pst(\bm{s}|\bm{u})$.
	  
    \item For all $\bm{s}$ and $\bm{u}$,
  	  $\frac{\delta(\bm{s}|\bm{u})}{\pst(\bm{s}|\bm{u})}$,
  	  $\frac{\delta^{l}(\bm{s}|\bm{u})}{\pst(\bm{s}|\bm{u})}$ and
  	  $\frac{\delta^{l,m}(\bm{s}|\bm{u})}{\pst(\bm{s}|\bm{u})}$ are
  	  finite where
  	  $\delta^{l}(\bm{s}|\bm{u}):=\parder{s_l}\delta(\bm{s}|\bm{u})$
  	  and $\delta^{l,m}(\bm{s}|\bm{u}):=\frac{\partial^2}{\partial
  	  s_l\partial s_m}\delta(\bm{s}|\bm{u})$.
	  
    \item The conditional density of $\bm{x}$ given $\bm{u}$ is
  	  universally approximated with an invertible feature extractor
  	  $\bm{h}(\bm{x})=(h_1(\bm{x}),\dots,h_{\dx}(\bm{x}))$ as
  	  \begin{align}
  	   \log\frac{p(\bm{x}|\bm{u})}{c(\bm{x})e(\bm{u})}
  	   =\sum_{i=1}^{\dx}\psi(h_i(\bm{x}),\bm{u}),
  	   \label{univ-approx-general}
  	  \end{align}
  	  where $\psi$, $c$ and $e$ are some functions.
   \end{enumerate}
   Then, under the contaminated density
   model~\eqref{source-contamination-model}, the following holds at an
   arbitrary fixed point $\bm{u}_0$:
   \begin{align}
    &\bm{M}(\bm{s})
    \bm{w}(\bm{s},\bm{u},\bm{u}_0)
    =\widetilde{\bm{M}}(\bm{s})
    \left(
    \begin{array}{c}
     \frac{\partial^2
     \{\log p(\bm{s}|\bm{u})-\log p(\bm{s}|\bm{u}_0)\}}{\partial s_1
     \partial s_2}\\
     \vdots\\
     \frac{\partial^2
      \{\log p(\bm{s}|\bm{u})-\log p(\bm{s}|\bm{u}_0)\}}{\partial s_{\dx}
      \partial s_{\dx-1}}
    \end{array}\right),
    \label{equation-M}
   \end{align}
   where $\bm{w}(\bm{s},\bm{u},\bm{u}_0)$ is the $2\dx$-dimensional
   vector of $\frac{\partial^2\left\{\log p(\bm{s}|\bm{u})-\log
   p(\bm{s}|\bm{u}_0)\right\}}{\partial s_i^2}$ and
   $\frac{\partial\left\{\log p(\bm{s}|\bm{u})-\log
   p(\bm{s}|\bm{u}_0)\right\}}{\partial s_i}$ for $i=1,\dots,\dx$.
   Moreover, when $\epsilon(\bm{u})$ is sufficiently small,
   $\frac{\partial^2 \log p(\bm{s}|\bm{u})}{\partial s_l \partial s_m}$
   for $l\neq m$ in the right-hand side of~\eqref{equation-M} equals to
   \begin{align}
    &\epsilon(\bm{u})\left\{
    \qsta{l}(s_l|\bm{u})\frac{\delta^{m}(\bm{s}|\bm{u})}{\pst(\bm{s}|\bm{u})}
    +\qsta{m}(s_m|\bm{u})
    \frac{\delta^{l}(\bm{s}|\bm{u})}{\pst(\bm{s}|\bm{u})}
    -\qsta{l}(s_l|\bm{u})\qsta{m}(s_m|\bm{u})\frac{\delta(\bm{s}|\bm{u})}
    {\pst(\bm{s}|\bm{u})}
    -\frac{\delta^{l,m}(\bm{s}|\bm{u})}{\pst(\bm{s}|\bm{u})}\right\}
    +O(\epsilon(\bm{u})^2), \label{defi-r}
   \end{align}   
   where $\qsta{l}(s_l|\bm{u}):=\parder{s_l}\qst(s_l|\bm{u})$.
  \end{theorem}
  The proof is given in Section~\ref{theo-general-case-proof}. First of
  all, in the case of no outliers, Theorem~\ref{theo-general-case}
  essentially recovers Theorem~1 in~\citet{hyvarinen2018nonlinear}: When
  $\epu=0$ for all $\bm{u}$, $\frac{\partial^2 \log
  p(\bm{s}|\bm{u})}{\partial s_l \partial s_m}=0$ for $l\neq m$ and thus
  $\bm{M}(\bm{s})\bm{w}(\bm{s},\bm{u},\bm{u}_0)=\bm{O}$, which leads to
  $\bm{M}(\bm{s})=\bm{O}$ (i.e., Eq.\eqref{general-condition-matrix})
  with an additional assumption that there exist $2\dx+1$ points,
  $\bm{u}_0, \bm{u}_1,\dots,\bm{u}_{2\dx}$, such that
  $\bm{w}(\bm{s},\bm{u}_1,\bm{u}_0),\dots,\bm{w}(\bm{s},\bm{u}_{2\dx},\bm{u}_0)$
  are linear independent.\footnote{This assumption is called
  \emph{Assumption of Variability} in~\citet{hyvarinen2018nonlinear},
  which implies that the conditional density $\pst(\bm{s}|\bm{u})$ is
  sufficiently complex and diverse.}

  Again, in the case of no outliers, \citet{hyvarinen2018nonlinear}
  asymptotically satisfies the universal approximation assumption~(B4)
  by applying logistic regression to the binary classification
  problem~\eqref{binary-classification} where the log-odds ratio,
  $\log\frac{p(\bm{x},\bm{u})}{p(\bm{x})p(\bm{u})}$, is approximated as
  the right-hand side of~\eqref{univ-approx-general} using neural
  networks.  Assumption~(B4) is a more general expression than the odds
  ratio, and indicates that in order to perform nonlinear ICA, it is
  sufficient that the numerator is the conditional density
  $p(\bm{s}|\bm{u})$ or joint density $p(\bm{s},\bm{u})$ up to the
  product of nonzero scalar functions of $\bm{s}$ and $\bm{u}$.
  
  However, when $\epu\neq 0$, the right-hand side on~\eqref{equation-M}
  is nonzero in general, and thus the source $\bm{s}$ cannot be
  recovered. In particular, $\frac{\partial^2 \log
  p(\bm{s}|\bm{u})}{\partial s_l \partial s_m}$ in~\eqref{equation-M} is
  the distortion factor, which can be expressed from the four ratios
  related to the outlier density $\delta(\bm{s}|\bm{u})$:
  $\frac{\delta^{m}(\bm{s}|\bm{u})}{\pst(\bm{s}|\bm{u})}$,
  $\frac{\delta^{l}(\bm{s}|\bm{u})}{\pst(\bm{s}|\bm{u})}$,
  $\frac{\delta(\bm{s}|\bm{u})}{\pst(\bm{s}|\bm{u})}$ and
  $\frac{\delta^{l,m}(\bm{s}|\bm{u})}{\pst(\bm{s}|\bm{u})}$.  These
  ratios can be very large when smooth $\delta(\bm{s}|\bm{u})$ lies on
  the tails of $\pst(\bm{s}|\bm{u})$.
  
  \subsection{Proof of Theorem~\ref{theo-general-case}}  
  \label{theo-general-case-proof}
  \begin{proof}
   We first obtain the expression of $\log p(\bm{x}|\bm{u})$ by the
   change of variables from $\log p(\bm{s}|\bm{u})$ through the data
   generative model~\eqref{ICA-model} as
   \begin{align*}
    \log p(\bm{x}|\bm{u})
    &=\log p(\bm{g}(\bm{x})|\bm{u})+\log|\det \bm{J}_{\bm{g}}(\bm{x})|,
   \end{align*}
   where $\bm{g}:=\bm{f}^{-1}$, $\det$ denotes the determinant and
   $\bm{J}_{\bm{g}}(\bm{x})$ denotes the Jacobian.  Then, the universal
   approximation assumption~(B4) gives
   \begin{align*}
    \sum_{i=1}^{\dx} \psi(h_{i}(\bm{x}),\bm{u})
    =\log p(\bm{g}(\bm{x})|\bm{u})+\log|\det \bm{J}_{\bm{g}}(\bm{x})|.
   \end{align*}
   To remove the Jacobian term, we compute the differences of the above
   equation between $\bm{u}$ and a fixed point $\bm{u}_0$ as
   \begin{align*}
    \sum_{i=1}^{\dx} \bar{\psi}(h_{i}(\bm{x}),\bm{u},\bm{u}_0)
    =\log p(\bm{g}(\bm{x})|\bm{u})-\log p(\bm{g}(\bm{x})|\bm{u}_0).
   \end{align*}
   where
   \begin{align*}
    \bar{\psi}(h_{i}(\bm{x}),\bm{u},\bm{u}_0)
    &:=\psi(h_{i}(\bm{x}),\bm{u})-\psi(h_{i}(\bm{x}),\bm{u}_0).
   \end{align*}
   By the further change of variables $\bm{h}=\bm{h}(\bm{x})$,
   \begin{align}
    \sum_{i=1}^{\dx} \bar{\psi}(h_{i},\bm{u},\bm{u}_0)
    =\log p(\bm{s}(\bm{h})|\bm{u})-\log p(\bm{s}(\bm{h})|\bm{u}_0)
    =:\bar{\varphi}(\bm{s}(\bm{h}),\bm{u},\bm{u}_0),
    \label{app:univ-approx-general}
   \end{align}
   where $\bm{s}(\bm{h}):=\bm{g}(\bm{h}^{-1}(\bm{h}))$. In the
   following, we simply express $\bm{s}(\bm{h})$ as $\bm{s}$ whenever
   the dependency to $\bm{h}$ does not matter.
   
   Next, let us use the following notations:
   \begin{align*}
    \bar{\varphi}^l(\bm{s},\bm{u},\bm{u}_0) &:=\frac{\partial}{\partial
    s_l}\bar{\varphi}(\bm{s},\bm{u},\bm{u}_0), &
    \bar{\varphi}^{l,m}(\bm{s},\bm{u},\bm{u}_0)
    &:=\frac{\partial^2}{\partial s_l \partial
    s_m}\bar{\varphi}(\bm{s},\bm{u},\bm{u}_0)\\ s_l^k&:=
    \frac{\partial}{\partial h_k} s_l(\bm{h}),\qquad & s_l^{jk}&:=
    \frac{\partial^2}{\partial h_j \partial h_k} s_l(\bm{h})\\
    p^l(\bm{s}|\bm{u}) &:=\frac{\partial}{\partial s_l}p(\bm{s}|\bm{u}),
    & p^{l,m}(\bm{s}|\bm{u}) &:=\frac{\partial^2}{\partial s_l \partial
    s_m}p(\bm{s}|\bm{u})\\ \psta{l}(\bm{s}|\bm{u})
    &:=\frac{\partial}{\partial s_l}\pst(\bm{s}|\bm{u}),
    &\psta{l,m}(\bm{s}|\bm{u}) &:=\frac{\partial^2}{\partial s_l \partial
    s_m}\pst(\bm{s}|\bm{u})
   \end{align*}
   Taking the second-order partial derivative of the left-hand side
   on~\eqref{app:univ-approx-general} with respect to $h_j$ and $h_k$
   for $j\neq k$ yields
   \begin{align}
    \frac{\partial^2}{\partial h_j\partial h_k} \sum_{i=1}^{\dx}
    \bar{\psi}(h_{i},\bm{u},\bm{u}_0)=0.  \label{app:left-hand}
   \end{align}
   On the other hand, the second-order partial derivative of the
   right-hand side on~\eqref{app:univ-approx-general} can be expressed as
   \begin{align}
    \frac{\partial^2}{\partial h_j\partial h_k}
    \bar{\varphi}(\bm{s}(\bm{h}),\bm{u},\bm{u}_0)    
    &=\sum_{l=1}^d\left[
    \bar{\varphi}^{l,l}(\bm{s},\bm{u},\bm{u}_0)s_l^js_l^k
    +\bar{\varphi}^{l}(\bm{s},\bm{u},\bm{u}_0)s_l^{jk}\right]
    +\sum_{l=1}^{\dx}
    \sum_{\substack{m=1\\ m\neq l}}^{\dx}
    \bar{\varphi}^{l,m}(\bm{s},\bm{u},\bm{u}_0)s_l^js_m^k.\label{app:right-hand}
   \end{align}
   Equating~\eqref{app:left-hand} with~\eqref{app:right-hand}
   under~\eqref{app:univ-approx-general} gives the following equation:
   \begin{align}
    \sum_{l=1}^d\left[
    \bar{\varphi}^{l,l}(\bm{s},\bm{u},\bm{u}_0)s_l^js_l^k
    +\bar{\varphi}^{l}(\bm{s},\bm{u},\bm{u}_0)s_l^{jk}\right]
    =-\sum_{l=1}^{\dx}
    \sum_{\substack{m=1\\ m\neq l}}^{\dx}
    \bar{\varphi}^{l,m}(\bm{s},\bm{u},\bm{u}_0)s_l^js_m^k.
    \label{app:eqn-from-univ-approx}
   \end{align}
   Regarding $j=1,\dots,\dx$ and $k=1, \dots,\dx$, we collect all of
   $s_l^js_l^k$, $s_l^{jk}$ and $-s_l^js_m^k$ for $j\neq k$ and $l\neq
   m$ as $(\dx^2-\dx)$-dimensional vectors, $\bm{a}_l(\bm{s})$,
   $\bm{b}_l(\bm{s})$, and $\bm{c}_{l,m}(\bm{s})$, respectively. Then,
   \eqref{app:eqn-from-univ-approx} for all $j$ and $k$ (but $j\neq k$)
   can be expressed as
   \begin{align*}
    \sum_{l=1}^d\left[
    \bar{\varphi}^{l,l}(\bm{s},\bm{u},\bm{u}_0)\bm{a}_l(\bm{s})
    +\bar{\varphi}^{l}(\bm{s},\bm{u},\bm{u}_0)\bm{b}_l(\bm{s})\right]
    =\sum_{l=1}^{\dx} \sum_{\substack{m=1\\ m\neq l}}^{\dx}
    \bar{\varphi}^{l,m}(\bm{s},\bm{u},\bm{u}_0)\bm{c}_{l,m}(\bm{s}).
   \end{align*}
   Furthermore, the above equations for all $l$ and $m$ (but $l\neq m$)
   can be summarized as the following system of linear equations:
   \begin{align*}
    \bm{M}(\bm{s})\bm{w}(\bm{s},\bm{u},\bm{u}_0)=
    \widetilde{\bm{M}}(\bm{s})
    \bm{r}(\bm{s},\bm{u},\bm{u}_0),
   \end{align*}
   where
   $\bm{M}(\bm{s}):=(\bm{a}_1,\dots,\bm{a}_n,\bm{b}_1,\dots,\bm{b}_n)$,
   $\widetilde{\bm{M}}(\bm{s}):=(\bm{c}_{1,2},\bm{c}_{1,3},\dots,\bm{c}_{\dx,\dx-1})$,
   and $\bm{w}(\bm{s},\bm{u},\bm{u}_0)$ is the $2\dx$-dimensional vector
   of all $\bar{\varphi}^{l,l}(\bm{s},\bm{u},\bm{u}_0)$ and
   $\bar{\varphi}^{l}(\bm{s},\bm{u},\bm{u}_0)$, and
   $\bm{r}(\bm{s},\bm{u},\bm{u}_0)$ is the $(\dx^2-\dx)$-dimensional
   vector of all $\bar{\varphi}^{l,m}(\bm{s},\bm{u},\bm{u}_0)$ for
   $l\neq m$.
   
   Finally, we prove that $\frac{\partial^2}{\partial s_l\partial s_m}
   \log p(\bm{s}|\bm{u})$ for $l\neq m$ equals to~\eqref{defi-r}.  We
   first compute
   \begin{align}
    \frac{\partial^2}{\partial s_l\partial s_m}   
    \log p(\bm{s}|\bm{u})
    &=\frac{p^{l,m}(\bm{s}|\bm{u})p(\bm{s}|\bm{u})
    -p^{l}(\bm{s}|\bm{u})p^{m}(\bm{s}|\bm{u})}{p(\bm{s}|\bm{u})^2}
    \nonumber\\&=\frac{p^{l,m}(\bm{s}|\bm{u})p(\bm{s}|\bm{u})
    -p^{l}(\bm{s}|\bm{u})p^{m}(\bm{s}|\bm{u})}
    {\pst(\bm{s}|\bm{u})^2}(1+O(\epsilon(\bm{u})))\nonumber\\
    &=(1-\epsilon(\bm{u}))^2\frac{\psta{l,m}(\bm{s}|\bm{u})\pst(\bm{s}|\bm{u})
    -\psta{l}(\bm{s}|\bm{u})\psta{m}(\bm{s}|\bm{u})}{\pst(\bm{s}|\bm{u})^2}\nonumber\\
    &\quad+\epsilon(\bm{u})\frac{\psta{l}(\bm{s}|\bm{u})\delta^{m}(\bm{s}|\bm{u})
    +\psta{m}(\bm{s}|\bm{u})\delta^{l}(\bm{s}|\bm{u})
    -\psta{l,m}(\bm{s}|\bm{u})\delta(\bm{s}|\bm{u})
    -\pst(\bm{s}|\bm{u})\delta^{l,m}(\bm{s}|\bm{u})}{\pst(\bm{s}|\bm{u})^2}
    +O(\epu^2),
    \label{app:second-derivative}
   \end{align}
   where we used the contaminated density
   model~\eqref{source-contamination-model} on the third line and
   applied the following relation with a sufficiently small
   $\epsilon(\bm{u})$ on the second line:
   \begin{align*}
    \frac{1}{p(\bm{s}|\bm{u})^2}
    &=\frac{1}{\left\{(1-\epsilon(\bm{u}))\pst(\bm{s}|\bm{u})
    +\epsilon(\bm{u})\delta(\bm{s}|\bm{u})\right\}^2}
   =\frac{1}{\pst(\bm{s}|\bm{u})^2\left\{1
    +\epsilon(\bm{u})\left(\frac{\delta(\bm{s}|\bm{u})}
    {\pst(\bm{s}|\bm{u})}-1\right)\right\}^2}
    =\frac{1}{\pst(\bm{s}|\bm{u})^2}(1+O(\epsilon(\bm{u}))).
   \end{align*}
   When $l\neq m$, it can be verified under the conditional independence
   assumption~(B2) that the first term in~\eqref{app:second-derivative}
   equals to zero as
   \begin{align*}
    \frac{\psta{l,m}(\bm{s}|\bm{u})\pst(\bm{s}|\bm{u})
    -\psta{l}(\bm{s}|\bm{u})\psta{m}(\bm{s}|\bm{u})}{\pst(\bm{s}|\bm{u})^2} 
    =\frac{\partial^2}{\partial s_l\partial s_m} \log\pst(\bm{s}|\bm{u})
    =\frac{\partial}{\partial s_l}\qsta{m}(s_m|\bm{u})=0, 
   \end{align*}
   Thus, \eqref{app:second-derivative} becomes
   \begin{align*}
    &\frac{\partial^2}{\partial s_l\partial s_m}   
    \log p(\bm{s}|\bm{u})\nonumber\\
    &=
    \epsilon(\bm{u})\left\{
    \qsta{l}(s_l|\bm{u})\frac{\delta^{m}(\bm{s}|\bm{u})}{\pst(\bm{s}|\bm{u})}
    +\qsta{m}(s_m|\bm{u})\frac{\delta^{l}(\bm{s}|\bm{u})}{\pst(\bm{s}|\bm{u})}
    -\qsta{l}(s_l|\bm{u})\qsta{m}(s_m|\bm{u})\frac{\delta(\bm{s}|\bm{u})}{\pst(\bm{s}|\bm{u})}
    -\frac{\delta^{l,m}(\bm{s}|\bm{u})}{\pst(\bm{s}|\bm{u})}\right\}
    +O(\epsilon(\bm{u})^2),
   \end{align*}
   where we used
   \begin{align*}
    \frac{\psta{l}(\bm{s}|\bm{u})}{\pst(\bm{s}|\bm{u})}&=
    \parder{s_l}\log\pst(\bm{s}|\bm{u})=\qsta{l}(s_l|\bm{u})\\
    \frac{\psta{l,m}(\bm{s}|\bm{u})}{\pst(\bm{s}|\bm{u})}
    &=\frac{\partial^2}{\partial s_l\partial s_m}
    \log\pst(\bm{s}|\bm{u})    
    +\frac{\psta{l}(\bm{s}|\bm{u})\psta{m}(\bm{s}|\bm{u})}{\pst(\bm{s}|\bm{u})^2}
    =\qsta{l}(s_l|\bm{u})\qsta{m}(s_m|\bm{u}),
   \end{align*}   
   under the conditional independence assumption~(B2). Thus, the proof
   is completed.
  \end{proof}
 \section{Proof of Theorem~\ref{theo:robustness}}
 \label{app:robustness}
  \subsection{Derivation of~\eqref{gamma-decomposition}}
  With $p(y,\bm{x},\bm{u})=p(\bm{x},\bm{u}|y)p(y)$, we have
  \begin{align}
   d_{\gamma}(p(y|\bm{x},\bm{u}), r(\bm{x},\bm{u});
   p(\bm{x},\bm{u}))
   &:=-\frac{1}{\gamma}\log \iint \sum_{y=0}^{1}
   \left\{\frac{r(\bm{x},\bm{u})^{y(\gamma+1)}}{1+r(\bm{x},\bm{u})^{\gamma+1}}
   \right\}^{\frac{\gamma}{\gamma+1}}p(y,\bm{x},\bm{u})
  \intd\bm{x}\intd\bm{u}\nonumber\\
   &:=-\frac{1}{\gamma}\log \left[
   \iint \left\{\frac{r(\bm{x},\bm{u})^{\gamma+1}}
   {1+r(\bm{x},\bm{u})^{\gamma+1}}\right\}^{\frac{\gamma}{\gamma+1}}
   p(\bm{x},\bm{u}|y=1)p(y=1)\intd\bm{x}\intd\bm{u}\right.\nonumber\\
   &\left.\qquad\qquad\qquad+\iint
   \left\{\frac{1}{1+r(\bm{x},\bm{u})^{\gamma+1}}\right\}^{\frac{\gamma}{\gamma+1}}
   p(\bm{x},\bm{u}|y=0)p(y=0)
   \intd\bm{x}\intd\bm{u}\right]\nonumber\\
   &:=-\frac{1}{\gamma}\log \left[
   \iint \left\{\frac{r(\bm{x},\bm{u})^{\gamma+1}}
   {1+r(\bm{x},\bm{u})^{\gamma+1}}\right\}^{\frac{\gamma}{\gamma+1}}
   p(\bm{x},\bm{u})p(y=1)\intd\bm{x}\intd\bm{u}\right.\nonumber\\
   &\left.\qquad\qquad\qquad +\iint
   \left\{\frac{1}{1+r(\bm{x},\bm{u})^{\gamma+1}}\right\}^{\frac{\gamma}{\gamma+1}}
   p(\bm{x})p(\bm{u})p(y=0)\intd\bm{x}\intd\bm{u}\right],
   \label{org-gamma}
  \end{align}
  where $p(\bm{x},\bm{u}|y=1)=p(\bm{x},\bm{u})$, and
  $p(\bm{x},\bm{u}|y=0)=p(\bm{x})p(\bm{u})$. Under the outlier model, the
  joint density can be expressed as
   \begin{align*}
    p(\bm{x},\bm{u})&=p(\bm{x}|\bm{u})p(\bm{u})
    =(1-\epu)\pst(\bm{x},\bm{u})
    +\epu\delta(\bm{x},\bm{u}).
   \end{align*}
   Then, the first term inside the
   logarithm~\eqref{org-gamma} can be written as
   \begin{align}
   &\iint \left\{\frac{r(\bm{x},\bm{u})^{\gamma+1}}
    {1+r(\bm{x},\bm{u})^{\gamma+1}}\right\}^{\frac{\gamma}{\gamma+1}}p(\bm{x},\bm{u})\intd\bm{x}\intd\bm{u}\nonumber\\
   &=
    \iint\left\{\frac{r(\bm{x},\bm{u})^{\gamma+1}}
    {1+r(\bm{x},\bm{u})^{\gamma+1}}\right\}^{\frac{\gamma}{\gamma+1}}
    (1-\epu)\pst(\bm{x},\bm{u})\intd\bm{x}\intd\bm{u}
    +\iint\left\{\frac{r(\bm{x},\bm{u})^{\gamma+1}}
    {1+r(\bm{x},\bm{u})^{\gamma+1}}\right\}^{\frac{\gamma}{\gamma+1}}
    \epu\delta(\bm{x},\bm{u})\intd\bm{x}\intd\bm{u}\label{log-first-term}
   \end{align}
   Finally, substituting~\eqref{log-first-term} into~\eqref{org-gamma}
   yields
   \begin{align*}
    &d_{\gamma}(p(y|\bm{x},\bm{u}), r(\bm{x},\bm{u}); p(\bm{x},\bm{u}))\\
    &:=-\frac{1}{\gamma}\log \left[
    \frac{1}{2}\iint \left\{\frac{r(\bm{x},\bm{u})^{\gamma+1}}
    {1+r(\bm{x},\bm{u})^{\gamma+1}}\right\}^{\frac{\gamma}{\gamma+1}}
    (1-\epu)\pst(\bm{x},\bm{u})\intd\bm{x}\intd\bm{u}
    +\frac{1}{2}\iint
    \left\{\frac{1}{1+r(\bm{x},\bm{u})^{\gamma+1}}\right\}^{\frac{\gamma}{\gamma+1}}
    p(\bm{x})p(\bm{u})\intd\bm{x}\intd\bm{u}\right.\\
    &\left.\qquad\qquad\qquad +\frac{1}{2}\iint\left\{\frac{r(\bm{x},\bm{u})^{\gamma+1}}
    {1+r(\bm{x},\bm{u})^{\gamma+1}}\right\}^{\frac{\gamma}{\gamma+1}}
    \epu\delta(\bm{x},\bm{u})\intd\bm{x}\intd\bm{u}\right]\\
    &=J[r(\bm{x},\bm{u});(1-\epu)\pst(\bm{x},\bm{u}), p(\bm{x})p(\bm{u})]
    +O(\nu),
   \end{align*}
   where we applied the relation $\log(y+z)=\log(y) +O(z)$ with
   sufficiently small $z$.
   \subsection{Proof of the minimizer~\eqref{gamma-minimizer}}
   \subsubsection{Preliminaries}
   We use the following results in the main proof, which are derived
   from the Taylor expansion:
   \begin{align}
    \left(\frac{1}{(r+\eta\phi)^{\gamma+1}+1}\right)^{\frac{\gamma}{\gamma+1}}
    &=\left(\frac{1}{r^{\gamma+1}+1}\right)^{\frac{\gamma}{\gamma+1}}
    -\eta\frac{\gamma r^{\gamma}\phi}
    {\left(r^{\gamma+1}+1\right)^{\frac{2\gamma+1}{\gamma+1}}}
    +\frac{\eta^2}{2}\frac{\{\gamma(\gamma+1)r^{2\gamma}-\gamma^2r^{\gamma-1}\}\phi^2}
    {\left(r^{\gamma+1}+1\right)^{\frac{3\gamma+1}{\gamma+1}}}+O(\eta^3)
    \label{taylor1}\\
    \left(\frac{(r+\eta\phi)^{\gamma+1}}
    {(r+\eta\phi)^{\gamma+1}+1}\right)^{\frac{\gamma}{\gamma+1}}
    &=\left(\frac{r^{\gamma+1}}{r^{\gamma+1}+1}\right)^{\frac{\gamma}{\gamma+1}}
    +\eta\frac{\gamma r^{\gamma-1}\phi}
    {\left(r^{\gamma+1}+1\right)^{\frac{2\gamma+1}{\gamma+1}}}
    -\frac{\eta^2}{2}\frac{\{\gamma(\gamma+2)r^{2\gamma-1}
    -\gamma(\gamma-1)r^{\gamma-2}\}\phi^2}
    {\left(r^{\gamma+1}+1\right)^{\frac{3\gamma+2}{\gamma+1}}}+O(\eta^3).
    \label{taylor2}   
   \end{align}
   \subsubsection{Main proof}
   \begin{proof}
    Let us define $\tilde{J}[r]:=\exp(-\gamma
    J[r(\bm{x},\bm{u});(1-\epu)\pst(\bm{x},\bm{u}),
    p(\bm{x})p(\bm{u})])$, and then we show a maximizer of $\tilde{J}[r]$
    alternative to a minimizer of
    $J[r(\bm{x},\bm{u});(1-\epu)\pst(\bm{x},\bm{u}),
    p(\bm{x})p(\bm{u})])$. For $\eta>0$ and a perturbation $\phi$, with
    \eqref{taylor1} and~\eqref{taylor2}, we have
    \begin{align}
     \tilde{J}[r+\eta\phi]&=\tilde{J}[r]
     +\frac{\eta}{2}\int\int\left[
     \frac{\gamma r(\bm{x},\bm{u})^{\gamma-1}
     \phi(\bm{x},\bm{u})\{(1-\epu)\pst(\bm{x},\bm{u})
     -r(\bm{x},\bm{u})p(\bm{x})p(\bm{u})\}}
     {\left(r(\bm{x},\bm{u})^{\gamma+1}+1\right)^{\frac{2\gamma+1}{\gamma+1}}}\right]
     \intd\bm{x}\intd\bm{u} \nonumber \\
     &+\frac{\eta^2}{4}\iint\left[
     \frac{\left\{\gamma(\gamma+1)r(\bm{x},\bm{u})^{2\gamma}
    -\gamma^2r(\bm{x},\bm{u})^{\gamma-1}\right\}
     p(\bm{x})p(\bm{u})\phi(\bm{x},\bm{u})^2}    
     {\left(r(\bm{x},\bm{u})^{\gamma+1}+1\right)^{\frac{3\gamma+2}{\gamma+1}}}\right.
     \nonumber\\  &\left.
    -\frac{\left\{\gamma(\gamma+2)r(\bm{x},\bm{u})^{2\gamma-1}
     -\gamma(\gamma-1)r(\bm{x},\bm{u})^{\gamma-2}\right\}(1-\epu)\pst(\bm{x},\bm{u})
     \phi(\bm{x},\bm{u})^2}
    {\left(r(\bm{x},\bm{u})^{\gamma+1}+1\right)^{\frac{3\gamma+2}{\gamma+1}}}\right]
     \intd\bm{x}\intd\bm{u}+O(\eta^3).
     \label{perturbation}
    \end{align}
    
    The optimality condition is satisfied when the term of order $\eta$
    on the right-hand side of~\eqref{perturbation} equals to zero for
    arbitrary $\phi$. Thus, an optimizer satisfies the following
    equation:
    \begin{align*}
     (1-\epu)\pst(\bm{x},\bm{u})-r(\bm{x},\bm{u})p(\bm{x})p(\bm{u})=0.
    \end{align*}
    Then, we obtain the optimizer as
    \begin{align*}
     r^{\star}(\bm{x},\bm{u})=\frac{(1-\epu)\pst(\bm{x},\bm{u})}{p(\bm{x})p(\bm{u})}
     =\frac{(1-\epu)\pst(\bm{x}|\bm{u})}{p(\bm{x})}.
    \end{align*}
    To investigate if $r^{\star}$ is a maximizer of $J[r]$, we compute
    the term of order $\eta^2$ on the right-hand side
    of~\eqref{perturbation} at $r=r^{\star}$ as
    \begin{align*}
     -\iint\left[
     \frac{\gamma\left\{r^{\star}(\bm{x},\bm{u})^{2\gamma}
     +r^{\star}(\bm{x},\bm{u})^{\gamma-1}\right\}
     p(\bm{x})p(\bm{u})\phi(\bm{x},\bm{u})^2}   
     {\left(r^{\star}(\bm{x},\bm{u})^{\gamma+1}+1\right)^{\frac{3\gamma+1}{\gamma+1}}}
     \right] \intd\bm{x}\intd\bm{u},
    \end{align*}
    where we used the relation,
    $(1-\epu)\pst(\bm{x},\bm{u})=r^{\star}(\bm{x},\bm{u})p(\bm{x})p(\bm{u})$. This
    shows that the term of order $\eta^2$ is negative for any choices of
    $\phi$. Thus, $r^{\star}$ is a maximizer, and the proof is
    completed.
   \end{proof}     
 
 \section{Proof of Proposition~\ref{prop:nu}}
 \label{app:proof-proposition}
 \begin{proof}
 In the neighborhood of
  $r^{\star}(\bm{x},\bm{u})=\frac{(1-\epu)\pst(\bm{x}|\bm{u})}{p(\bm{x})}$,
 we can obtain
 \begin{align*}
  \nu&=\iint_{\mathcal{S},\mathcal{U}}
  \left\{\frac{r(\bm{x},\bm{u})^{\gamma+1}}
  {1+r(\bm{x},\bm{u})^{\gamma+1}}\right\}^{\frac{\gamma}{\gamma+1}}
  \epu\delta(\bm{x},\bm{u})\intd\bm{x}\intd\bm{u}\\ 
  &\leq\iint_{\mathcal{S},\mathcal{U}}
  \left\{\frac{\pst(\bm{x}|\bm{u})^{\gamma+1}}
  {\{(1-\epu)\pst(\bm{x}|\bm{u})\}^{\gamma+1}+p(\bm{x})^{\gamma+1}}
  \right\}^{\frac{\gamma}{\gamma+1}}
  \delta(\bm{x}|\bm{u})\intd\bm{x} (1-\epu)^{\gamma}\epu p(\bm{u})\intd\bm{u}
  +O\left(\sup_{\bm{x},\bm{u}}|r(\bm{x},\bm{u})-r^{\star}(\bm{x},\bm{u})|\right).
 \end{align*}
 Next, we perform the change of variables from $\bm{x}$ to $\bm{s}$
 based on the generative model~\eqref{ICA-model} and have
 \begin{align*}
  \nu&\leq\int_{\mathcal{U}}
  \left[\int_{\mathcal{S}}\left\{\frac{\pst(\bm{s}|\bm{u})^{\gamma+1}}
  {\{(1-\epu)\pst(\bm{s}|\bm{u})\}^{\gamma+1}+p(\bm{s})^{\gamma+1}}
  \right\}^{\frac{\gamma}{\gamma+1}}
  \delta(\bm{s}|\bm{u})\intd\bm{s}\right]g_{\gamma}(\bm{u})\intd\bm{u}
  +O\left(\sup_{\bm{x},\bm{u}}|r(\bm{x},\bm{u})-r^{\star}(\bm{x},\bm{u})|\right),
 \end{align*}
 where $g_{\gamma}(\bm{u}):=(1-\epu)^{\gamma}\epu p(\bm{u})$.  We note
 that the ratio inside the integral does not depend on the partition
 function, and thus the change of variables is possible easily.
  
 Under the assumption that
 $\mathcal{S}^{\pst}_{\bm{u}}\cap\mathcal{S}^{\delta}_{\bm{u}}=\emptyset$,
 we can easily conform that for $\gamma>0$,
 \begin{align*}
  \int_{\mathcal{S}}\left\{\frac{\pst(\bm{s}|\bm{u})^{\gamma+1}}
  {\{(1-\epu)\pst(\bm{s}|\bm{u})\}^{\gamma+1}+p(\bm{s})^{\gamma+1}}
  \right\}^{\frac{\gamma}{\gamma+1}}
  \delta(\bm{s}|\bm{u})\intd\bm{s}=0.
  \end{align*}
 The above equation yields
 \begin{align*}
  \nu&\leq
  O\left(\sup_{\bm{x},\bm{u}}|r(\bm{x},\bm{u})-r^{\star}(\bm{x},\bm{u})|\right).
 \end{align*}
 Thus, the proof is completed.  
 \end{proof} 
   
 \section{Avoiding the influence from $\delta(\bm{s}|\bm{u})/\pst(\bm{s}|\bm{u})$}
 \label{app:neighbor-inequality}
 Section~\ref{sec:outliers} already indicated that one of the main
 factors to hamper estimation in nonlinear ICA might be the large
 density ratio $\delta(\bm{s}|\bm{u})/\pst(\bm{s}|\bm{u})$.  The
 following inequality derived in
 Section~\ref{app-sub:neighbor-inequality} reveals its connection:
 \begin{align}    
  \nu&\leq \frac{1}{C_{\gamma}}
  \int\underbrace{\left[\int \pst(\bm{s}|\bm{u})^{\gamma+1}
  \frac{\delta(\bm{s}|\bm{u})}{\pst(\bm{s}|\bm{u})}
  \intd\bm{s}\right]}_{(*)} g_{\gamma}(\bm{u}) \intd\bm{u}
  +O(\sup_{\bm{x},\bm{u}}|r(\bm{x},\bm{u})-r^{\star}(\bm{x},\bm{u})|),
  \label{neighborhood}   
 \end{align}
 where $g_{\gamma}(\bm{u}):=(1-\epu)^{\gamma}\epu p(\bm{u})$ and
 $C_{\gamma}:=\inf_{\bm{s},\bm{u}}[\{(1-\epu)\pst(\bm{s}|\bm{u})\}^{\gamma+1}
 +p(\bm{s})^{\gamma+1}]^{\gamma/(\gamma+1)}$ is assumed to be
 nonzero. Integral~$(*)$ implies that using a larger $\gamma$ would
 reduce the influence of the density ratio
 $\delta(\bm{s}|\bm{u})/\pst(\bm{s}|\bm{u})$ by multiplying it by the
 density power $\pst(\bm{s}|\bm{u})^{\gamma+1}$ in the neighborhood of
 $r^{\star}(\bm{x},\bm{u})$. In contrast, in the limit of $\gamma=0$,
 the integral~$(*)$ is constant, and therefore logistic regression might
 not be able to avoid a strong influence from
 $\delta(\bm{s}|\bm{u})/\pst(\bm{s}|\bm{u})$.
  \subsection{Derivation of Inequality~\eqref{neighborhood}}
  \label{app-sub:neighbor-inequality}
  In the neighborhood of
  $r^{\star}(\bm{x},\bm{u})=\frac{(1-\epu)\pst(\bm{x}|\bm{u})}{p(\bm{x})}$,
  we obtain
  \begin{align*}
  \nu&=\iint\left\{\frac{r(\bm{x},\bm{u})^{\gamma+1}}
  {1+r(\bm{x},\bm{u})^{\gamma+1}}\right\}^{\frac{\gamma}{\gamma+1}}
  \epu\delta(\bm{x},\bm{u})\intd\bm{x}\intd\bm{u}\\
   &\leq\iint\left\{\frac{\{(1-\epu)\pst(\bm{x},\bm{u})\}^{\gamma+1}}
   {\{(1-\epu)\pst(\bm{x},\bm{u})\}^{\gamma+1}+\{p(\bm{x})p(\bm{u})\}^{\gamma+1}}
   \right\}^{\frac{\gamma}{\gamma+1}}\delta(\bm{x})p(\bm{u})\intd\bm{x}\intd\bm{u}
  +O\left(\sup_{\bm{x},\bm{u}}|r(\bm{x},\bm{u})-r^{\star}(\bm{x},\bm{u})|\right)\\
  &=\iint\left\{\frac{\pst(\bm{x}|\bm{u})^{\gamma+1}}
  {\{(1-\epu)\pst(\bm{x}|\bm{u})\}^{\gamma+1}+p(\bm{x})^{\gamma+1}}
  \right\}^{\frac{\gamma}{\gamma+1}} \delta(\bm{x}|\bm{u})\intd\bm{x}
  (1-\epu)^{\gamma}\epu p(\bm{u})\intd\bm{u}
  +O\left(\sup_{\bm{x},\bm{u}}|r(\bm{x},\bm{u})-r^{\star}(\bm{x},\bm{u})|\right).
  \end{align*}
  Next, we perform the change of variables from $\bm{x}$ to $\bm{s}$
  based on the generative model~\eqref{ICA-model} and have
  \begin{align*}
   \nu&\leq
   \int\left[\int\left\{\frac{\pst(\bm{s}|\bm{u})^{\gamma+1}}
   {\{(1-\epu)\pst(\bm{s}|\bm{u})\}^{\gamma+1}+p(\bm{s})^{\gamma+1}}
   \right\}^{\frac{\gamma}{\gamma+1}}
   \delta(\bm{s}|\bm{u})\intd\bm{s}\right]g_{\gamma}(\bm{u})\intd\bm{u}
   +O\left(\sup_{\bm{x},\bm{u}}|r(\bm{x},\bm{u})-r^{\star}(\bm{x},\bm{u})|\right),
  \end{align*}
  where $g_{\gamma}(\bm{u}):=(1-\epu)^{\gamma}\epu p(\bm{u})$.  Thus,
  \begin{align*}
   \nu&\leq 
   \frac{1}{C_{\gamma}}
   \int\left[\int \pst(\bm{s}|\bm{u})^{\gamma}
   \delta(\bm{s}|\bm{u})\intd\bm{s}\right]g_{\gamma}(\bm{u})\intd\bm{u}
   +O\left(\sup_{\bm{x},\bm{u}}|r(\bm{x},\bm{u})-r^{\star}(\bm{x},\bm{u})|\right)\\
   &=
   \frac{1}{C_{\gamma}}
   \int\left[\int \pst(\bm{s}|\bm{u})^{\gamma+1}
   \frac{\delta(\bm{s}|\bm{u})}{\pst(\bm{s}|\bm{u})}
   \intd\bm{s}\right]g_{\gamma}(\bm{u})\intd\bm{u}
   +O\left(\sup_{\bm{x},\bm{u}}|r(\bm{x},\bm{u})-r^{\star}(\bm{x},\bm{u})|\right),
  \end{align*}  
  where $C_{\gamma}:=\inf_{\bm{s},\bm{u}}
  |\{(1-\epu)\pst(\bm{s}|\bm{u})\}^{\gamma+1}+p(\bm{s})^{\gamma+1}|$.
 \section{Proof of Proposition~\ref{proposition.if1}}
 \label{app:proof-robust}

 \begin{proof}
  We first derive the following lemma, which shows the form of the
  influence function:
  \begin{lemma}
   \label{lemma:if} The influence function ${\rm IF}(\bx,\bu)$ of
   $\vhtheta$ under the $\gamma$-cross entropy~\eqref{binary} satisfies
   \begin{align*}
    C_{\vhtheta} {\rm IF}(\bx,\bu)=& \iint
    \left\{A_{\vhtheta}(\bm{x},\bm{u})\bar{\delta}_{(\bx,\bu)}(\bm{x},\bm{u})
    -B_{\vhtheta}(\bm{x},\bm{u})\left(
    \pst(\bm{x})\bar{\delta}_{\bu}(\bm{u})+\pst(\bm{u})\bar{\delta}_{\bx}(\bm{x})
    \right)
    +B_{\vhtheta}(\bm{x},\bm{u})\pst(\bm{x})\pst(\bm{u})\right\}\intdx\intdu
    \\
    =& 
    A_{\vhtheta}(\bx,\bu)
    -\int B_{\vhtheta}(\bm{x},\bu)\pst(\bm{x})\intdx
    -\int B_{\vhtheta}(\bx,\bm{u})\pst(\bm{u})\intdu
    +\iint
    B_{\vhtheta}(\bm{x},\bm{u})\pst(\bm{x})\pst(\bm{u})\intdx\intdu,    
   \end{align*}
   where
   \begin{align*}
    L_{\vtheta}(\bm{x},\bm{u})&=\frac{1}{1+\tf},  \\
    S_{\vtheta}(\bm{x},\bm{u})&=\left\{
    L_{\vtheta}(\bm{x},\bm{u})(1-L_{\vtheta}(\bm{x},\bm{u}))
    \right\}^{\gamma/(1+\gamma)},  \\
   A_{\vtheta}(\bm{x},\bm{u})&=L_{\vtheta}(\bm{x},\bm{u})^{1/(1+\gamma)}
    S_{\vtheta}(\bm{x},\bm{u})\frac{\partial \log \tf}{\partial \vtheta},
    \nonumber  \\
    B_{\vtheta}(\bm{x},\bm{u})&=\left(1-L_{\vtheta}(\bm{x},\bm{u})\right)^{1/(1+\gamma)}
    S_{\vtheta}(\bm{x},\bm{u})
    \frac{\partial \log \tf }{\partial \vtheta},   \nonumber\\ 
    C_{\vtheta}&=\iint\left\{
    \frac{\partial}{\partial\vtheta}A_{\vtheta}(\bm{x},\bm{u})^{\top}
    \pst(\bm{x},\bm{u}),
    -\frac{\partial}{\partial\vtheta}B_{\vtheta}(\bm{x},\bm{u})^{\top}
    \pst(\bm{x})\pst(\bm{u})\right\}\intdx\intdu.
    \label{if.matrix}
   \end{align*}
  \end{lemma}
  
  Let us prove Lemma~\ref{lemma:if}. $\vhtheta$ associated with the
  densities $\pst(\bm{x},\bm{u})$ and $\pst(\bm{x})\pst(\bm{u})$ is a
  solution of the following estimating equation:
  \begin{align}
   \bm{0}&=\left. 
   \parder{\bm{\theta}}d_{\gamma}(\pst(y|\bm{x},\bm{u}),
   r_{\bm{\theta}}(\bm{x},\bm{u});\pst(\bm{x},\bm{u}))  
   \right|_{\bm{\theta}=\vhtheta}  \\  
   &\propto \left[\iint\left\{
   \left(\frac{\tf}{1+\tf}\right)^{-1/(1+\gamma)}
   \frac{\partial_{\vtheta}\tf}{(1+\tf)^2}\pst(\bm{x},\bm{u})\right.\right.
   \nonumber \\ &  \left.\left. \left.\qquad 
   -\left(\frac{1}{1+\tf}\right)^{-1/(1+\gamma)}
   \frac{\partial_{\vtheta}\tf}{(1+\tf)^2}\pst(\bm{x})\pst(\bm{u})\right\} \intdx\intdu
   \right]\right|_{\bm{\theta}=\vhtheta},
   \label{estimating.target}
  \end{align}
  where $\partial_{\vtheta}:=\frac{\partial}{\partial \vtheta}$.  On the
  other hand, $\vhtheta_{\epsilon}$ is a solution of the the estimating
  equation with the contaminated distributions:
  \begin{align}
   \bm{0}&=\left.
   \parder{\bm{\theta}}d_{\gamma}(\bar{p}(y|\bm{x},\bm{u}),
   r_{\bm{\theta}}(\bm{x},\bm{u});\bar{p}(\bm{x},\bm{u}))   
   \right|_{\bm{\theta}=\vhtheta_{\epsilon}}\nonumber \\ 
   &\propto  \left[\iint\left\{
   \left(\frac{\tf}{1+\tf}\right)^{-1/(1+\gamma)}
   \frac{\partial_{\vtheta}\tf}{(1+\tf)^2}\bar{p}(\bm{x},\bm{u})
   \right.\right.  \nonumber \\ & \left.\left.\left.  \qquad-
   \left(\frac{1}{1+\tf}\right)^{-1/(1+\gamma)}
   \frac{\partial_{\vtheta}\tf}{(1+\tf)^2}\bar{p}(\bm{x})\bar{p}(\bm{u})
   \right\}\intdx\intdu\right]\right|_{\bm{\theta}=\vhtheta_{\epsilon}}.
   \label{app:estimating.eq}
  \end{align}
  Taylor series of~\eqref{app:estimating.eq} around $\vhtheta$ is given
  by
  \begin{align}
   0= \left.\frac{\partial d_{\gamma}}{\partial \vtheta}
   \right|_{\bm{\theta}=\vhtheta} 
   +\left.\frac{\partial d_{\gamma}}{\partial\vtheta \partial\vtheta^{\top}}
   \right|_{\bm{\theta}=\vhtheta}(\vhtheta_{\epsilon}-\vhtheta)
   +O(\|\vhtheta_{\epsilon}-\vhtheta\|^2). \label{Taylor}
  \end{align}
  Taking the limit of $\epsilon\to 0$ proves Lemma~\ref{lemma:if}
  using~\eqref{estimating.target}.  

  Since $L_{\vtheta}(\bm{x},\bm{u})^{1/(1+\gamma)}\leq 1$,
  we observe that
  \begin{align}
   \lim_{|r_{\vtheta}(\bm{x},\bm{u})|\to\infty}A_{\vtheta}(\bm{x},\bm{u})=
   \lim_{|r_{\vtheta}(\bm{x},\bm{u})|\to\infty}B_{\vtheta}(\bm{x},\bm{u})=0,
   \label{A-B-zero}
  \end{align}
  under Assumption \eqref{condition:if1}.  Eq.\eqref{A-B-zero} ensures
  that the influence function in Lemma~\ref{lemma:if} is bounded even
  when $r_{\vhtheta}(\bx,\bu)$, $r_{\vhtheta}(\bx,\bm{u})$ or
  $r_{\vhtheta}(\bm{x},\bu)$ diverge through the influence of the
  outliers $(\bx,\bu)$, {\it i.e.},
  \begin{align}
   \sup_{\bx,\bu}|{\rm IF}(\bx,\bu)|
   <\infty.
  \end{align}
  Thus, the proof is completed.
 \end{proof} 
 \section{Robust property in multiclass classification}
 \label{app:multi-robustness}
 We specifically assume that the following $\tilde{\nu}$ is sufficiently
 small:
 \begin{align*}
  \tilde{\nu}&:=
  \int\frac{\sum_{u}r(u,\bm{x})^{\gamma}\delta(\bm{x}|u)\epsilon(u)p(u)}
  {\left(\sum_{u'}r(u',\bm{x})^{\gamma+1}\right)^{\frac{\gamma}{\gamma+1}}}
  \intd\bm{x}.
 \end{align*}    
 Then, the outlier model~\eqref{source-contamination-model} decomposes
 the $\gamma$-cross entropy~\eqref{multiclass} into
 \begin{align}   
  &d_{\gamma}(p(u|\bm{x}),r(u,\bm{x});p(\bm{x}))\nonumber\\
  &=-\frac{1}{\gamma}\log\int\frac{\sum_{u}r(u,\bm{x})^{\gamma}p(u|\bm{x})}
  {\left(\sum_{u'}r(u',\bm{x})^{\gamma+1}\right)^{\frac{\gamma}{\gamma+1}}}
  p(\bm{x})\intd\bm{x}\nonumber\\
  &=-\frac{1}{\gamma}\log\left[
  \int\frac{\sum_{u}r(u,\bm{x})^{\gamma}(1-\epsilon(u))\pst(\bm{x}|u)p(u)}
  {\left(\sum_{u'}r(u',\bm{x})^{\gamma+1}\right)^{\frac{\gamma}{\gamma+1}}}
  \intd\bm{x}
  +\int\frac{\sum_{u}r(u,\bm{x})^{\gamma}\delta(\bm{x}|u)\epsilon(u)p(u)}
  {\left(\sum_{u'}r(u',\bm{x})^{\gamma+1}\right)^{\frac{\gamma}{\gamma+1}}}
  \intd\bm{x}\right]\nonumber\\
  &=d_{\gamma}(r(u,\bm{x}),\pst(\bm{x}|u);(1-\epsilon(u))p(u))
  +O(\tilde{\nu}),
  \label{entropy-decomposition-multiclass}
 \end{align}  
 where we employed $\log(y+z)=\log(y) +O(z)$ with sufficiently small $z$
 and $p(u|\bm{x})=\frac{p(\bm{x}|u)p(u)}{p(\bm{x})}
 =\frac{(1-\epsilon(u))\pst(\bm{x}|u)p(u)+\epsilon(u)\delta(\bm{x}|u)p(u)}{p(\bm{x})}$,
 and
 \begin{align*}
  d_{\gamma}(r(u,\bm{x}),\pst(\bm{x}|u);(1-\epsilon(u))p(u))
  &:=-\frac{1}{\gamma}\log\left[
  \sum_{u}\left\{\int\frac{r(u,\bm{x})^{\gamma}\pst(\bm{x}|u)}
  {\left(\sum_{u'}r(u',\bm{x})^{\gamma+1}\right)^{\frac{\gamma}{\gamma+1}}}
  \intd\bm{x}\right\}(1-\epsilon(u))p(u)\right].
 \end{align*}
 $d_{\gamma}(r(u,\bm{x}),\pst(\bm{x}|u);(1-\epsilon(u))p(u))$ is the
 $\gamma$-cross entropy to $\pst(\bm{x}|u)$ under the measure
 $(1-\epsilon(u))p(u)$, and thus is minimized at
 $r(u,\bm{x})=\pst(\bm{x}|u)$.  The minimizer is desirable in terms of
 the universal approximation assumptions (A4) and (B4) because it is the
 (noncontaminated) target density.

 Next, we discuss when $\tilde{\nu}$ is sufficiently small.  Following
 Section~\ref{app:proof-proposition}, in the neighborhood of
 $\pst(\bm{x}|u)$,
 \begin{align}
  \tilde{\nu}
  &\leq\sum_{u}\left\{
  \int\frac{\pst(\bm{x}|u)^{\gamma}\delta(\bm{x}|u)}
  {\left(\sum_{u'}\pst(\bm{x}|u')^{\gamma+1}\right)^{\frac{\gamma}{\gamma+1}}}
  \intd\bm{x}\right\}\epsilon(u) p(u)
  +O(\sup_{u,\bm{x}}|r(u,\bm{x})-\pst(\bm{x}|u)|)\nonumber\\
  &=\sum_{u}\left\{
  \int\frac{\pst(\bm{s}|u)^{\gamma}\delta(\bm{s}|u)}
  {\left(\sum_{u'}\pst(\bm{s}|u')^{\gamma+1}\right)^{\frac{\gamma}{\gamma+1}}}
  \intd\bm{s}\right\}\epsilon(u) p(u)
  +O(\sup_{u,\bm{x}}|r(u,\bm{x})-\pst(\bm{x}|u)|),
  \label{multiclass-upper-bound}
 \end{align}
 where we performed the change of variables from $\bm{x}$ to $\bm{s}$
 under the data generate model~\eqref{ICA-model}.  Under the same
 support assumptions of $\pst(\bm{s}|u)$ and $\delta(\bm{s}|u)$ as
 Section~\ref{app:proof-proposition}, we can make the same implication
 as Proposition~\ref{prop:nu}: $\nu$ can be sufficiently small in the
 neighborhood of $\pst(\bm{x}|u)$ when $\pst(\bm{s}|u)$ and
 $\delta(\bm{s}|u)$ are clearly separated. This clear separation
 possibly happens on a situation where $\delta(\bm{s}|u)$ lies on the
 tails of $\pst(\bm{s}|u)$ as in common outlier situations. Thus, the
 $\gamma$-cross entropy for multiclass classification would be also
 robust against outliers.
 \bibliography{../../../papers,../robustTCL_ricardo/FromRicardo/RobustTCL/ref}

\begin{thebibliography}{34}
\providecommand{\natexlab}[1]{#1}
\providecommand{\url}[1]{\texttt{#1}}
\expandafter\ifx\csname urlstyle\endcsname\relax
  \providecommand{\doi}[1]{doi: #1}\else
  \providecommand{\doi}{doi: \begingroup \urlstyle{rm}\Url}\fi

\bibitem[Amari(2016)]{amari2016information}
S.~Amari.
\newblock \emph{Information Geometry and Its Applications}.
\newblock Springer, 2016.

\bibitem[Basu et~al.(1998)Basu, Harris, Hjort, and Jones]{basu1998robust}
A.~Basu, I.~Harris, N.~Hjort, and M.~Jones.
\newblock Robust and efficient estimation by minimising a density power
  divergence.
\newblock \emph{Biometrika}, 85\penalty0 (3):\penalty0 549--559, 1998.

\bibitem[Chen et~al.(2013)Chen, Hung, Komori, Huang, and
  Eguchi]{chen2013robust}
P.~Chen, H.~Hung, O.~Komori, S.-Y. Huang, and S.~Eguchi.
\newblock Robust independent component analysis via minimum $\gamma
  $-divergence estimation.
\newblock \emph{{IEEE} Journal of Selected Topics in Signal Processing},
  7\penalty0 (4):\penalty0 614--624, 2013.

\bibitem[Cichocki and Amari(2010)]{cichocki2010families}
A.~Cichocki and S.~Amari.
\newblock Families of alpha-beta-and gamma-divergences: Flexible and robust
  measures of similarities.
\newblock \emph{Entropy}, 12\penalty0 (6):\penalty0 1532--1568, 2010.

\bibitem[Comon(1994)]{comon1994independent}
P.~Comon.
\newblock {Independent component analysis, a new concept?}
\newblock \emph{Signal Processing}, 36\penalty0 (3):\penalty0 287--314, 1994.

\bibitem[Fujisawa and Eguchi(2008)]{fujisawa2008robust}
H.~Fujisawa and S.~Eguchi.
\newblock Robust parameter estimation with a small bias against heavy
  contamination.
\newblock \emph{Journal of Multivariate Analysis}, 99\penalty0 (9):\penalty0
  2053--2081, 2008.

\bibitem[Goodfellow et~al.(2013)Goodfellow, Warde-Farley, Mirza, Courville, and
  Bengio]{goodfellow2013maxout}
I.~Goodfellow, D.~Warde-Farley, M.~Mirza, A.~Courville, and Y.~Bengio.
\newblock Maxout networks.
\newblock In \emph{Proceedings of the 30th International Conference on Machine
  Learning ({ICML})}, Proceedings of Machine Learning Research, pages
  1319--1327. PMLR, 2013.

\bibitem[Goodfellow et~al.(2014)Goodfellow, Pouget-Abadie, Mirza, Xu,
  Warde-Farley, Ozair, Courville, and Bengio]{goodfellow2014generative}
I.~Goodfellow, J.~Pouget-Abadie, M.~Mirza, B.~Xu, D.~Warde-Farley, S.~Ozair,
  A.~Courville, and Y.~Bengio.
\newblock Generative adversarial nets.
\newblock In \emph{Advances in neural information processing systems
  ({NeurIPS})}, pages 2672--2680, 2014.

\bibitem[Gretton et~al.(2005)Gretton, Bousquet, Smola, and
  Sch{\"o}lkopf]{gretton2005measuring}
A.~Gretton, O.~Bousquet, A.~Smola, and B.~Sch{\"o}lkopf.
\newblock Measuring statistical dependence with {H}ilbert-{S}chmidt norms.
\newblock In \emph{International conference on algorithmic learning theory},
  pages 63--77. Springer, 2005.

\bibitem[Gutmann and Hyv{\"a}rinen(2012)]{Gutmann2012a}
M.~Gutmann and A.~Hyv{\"a}rinen.
\newblock {N}oise-contrastive estimation of unnormalized statistical models,
  with applications to natural image statistics.
\newblock \emph{Journal of Machine Learning Research}, 13:\penalty0 307--361,
  2012.

\bibitem[Hampel et~al.(2011)Hampel, Ronchetti, Rousseeuw, and
  Stahel]{hampel2011robust}
F.~R. Hampel, E.~M. Ronchetti, P.~J. Rousseeuw, and W.~A. Stahel.
\newblock \emph{Robust statistics: the approach based on influence functions}.
\newblock John Wiley \& Sons, 2011.

\bibitem[Harmeling et~al.(2003)Harmeling, Ziehe, Kawanabe, and
  M{\"u}ller]{harmeling2003kernel}
S.~Harmeling, A.~Ziehe, M.~Kawanabe, and K.-R. M{\"u}ller.
\newblock Kernel-based nonlinear blind source separation.
\newblock \emph{Neural Computation}, 15\penalty0 (5):\penalty0 1089--1124,
  2003.

\bibitem[Hung et~al.(2018)Hung, Jou, and Huang]{hung2018robust}
H.~Hung, Z.-Y. Jou, and S.-Y. Huang.
\newblock Robust mislabel logistic regression without modeling mislabel
  probabilities.
\newblock \emph{Biometrics}, 74\penalty0 (1):\penalty0 145--154, 2018.

\bibitem[Hyv{\"a}rinen(1999)]{hyvarinen1999fast}
A.~Hyv{\"a}rinen.
\newblock Fast and robust fixed-point algorithms for independent component
  analysis.
\newblock \emph{IEEE Transactions on Neural Networks}, 10\penalty0
  (3):\penalty0 626--634, 1999.

\bibitem[Hyv{\"a}rinen and Morioka(2016)]{hyvarinen2016unsupervised}
A.~Hyv{\"a}rinen and H.~Morioka.
\newblock Unsupervised feature extraction by time-contrastive learning and
  nonlinear {ICA}.
\newblock In \emph{Advances in Neural Information Processing Systems
  ({NeurIPS})}, pages 3765--3773, 2016.

\bibitem[Hyv{\"a}rinen and Morioka(2017)]{pmlr-v54-hyvarinen17a}
A.~Hyv{\"a}rinen and H.~Morioka.
\newblock Nonlinear {ICA} of temporally dependent stationary sources.
\newblock In \emph{Proceedings of the 20th International Conference on
  Artificial Intelligence and Statistics ({AISTATS})}, volume~54, pages
  460--469. PMLR, 2017.

\bibitem[Hyv{\"a}rinen and Oja(2000)]{hyvarinen2000independent}
A.~Hyv{\"a}rinen and E.~Oja.
\newblock {Independent component analysis: algorithms and applications}.
\newblock \emph{Neural Networks}, 13\penalty0 (4--5):\penalty0 411--430, 2000.

\bibitem[Hyv{\"a}rinen and Pajunen(1999)]{hyvarinen1999nonlinear}
A.~Hyv{\"a}rinen and P.~Pajunen.
\newblock Nonlinear independent component analysis: Existence and uniqueness
  results.
\newblock \emph{Neural Networks}, 12\penalty0 (3):\penalty0 429--439, 1999.

\bibitem[Hyv{\"a}rinen et~al.(2019)Hyv{\"a}rinen, Sasaki, and
  Turner]{hyvarinen2018nonlinear}
A.~Hyv{\"a}rinen, H.~Sasaki, and R.~E. Turner.
\newblock Nonlinear {ICA} using auxiliary variables and generalized contrastive
  learning.
\newblock In \emph{Proceedings of the 22th International Conference on
  Artificial Intelligence and Statistics ({AISTATS})}, volume~89, pages
  859--868, 2019.

\bibitem[Kanamori and Fujisawa(2015)]{kanamori2015robust}
T.~Kanamori and H.~Fujisawa.
\newblock Robust estimation under heavy contamination using unnormalized
  models.
\newblock \emph{Biometrika}, 102\penalty0 (3):\penalty0 559--572, 2015.

\bibitem[Kawashima and Fujisawa(2018)]{kawashima2018difference}
T.~Kawashima and H.~Fujisawa.
\newblock On difference between two types of $\gamma $-divergence for
  regression.
\newblock \emph{arXiv:1805.06144}, 2018.

\bibitem[Kingma and Ba(2015)]{kingma2015adam}
D.~P. Kingma and J.~Ba.
\newblock {A}dam: {A} method for stochastic optimization.
\newblock In \emph{Proceedings of the 3rd International Conference on Learning
  Representations ({ICLR})}, pages 1--15, 2015.

\bibitem[Larsson et~al.(2017)Larsson, Maire, and
  Shakhnarovich]{larsson2017colorization}
G.~Larsson, M.~Maire, and G.~Shakhnarovich.
\newblock Colorization as a proxy task for visual understanding.
\newblock In \emph{Proceedings of the IEEE Conference on Computer Vision and
  Pattern Recognition ({CVPR})}, pages 6874--6883, 2017.

\bibitem[Locatello et~al.(2019)Locatello, Bauer, Lucic, Raetsch, Gelly,
  Sch{\"o}lkopf, and Bachem]{pmlr-v97-locatello19a}
F.~Locatello, S.~Bauer, M.~Lucic, G.~Raetsch, S.~Gelly, B.~Sch{\"o}lkopf, and
  O.~Bachem.
\newblock Challenging common assumptions in the unsupervised learning of
  disentangled representations.
\newblock In \emph{Proceedings of the 36th International Conference on Machine
  Learning ({ICML})}, volume~97 of \emph{Proceedings of Machine Learning
  Research}, pages 4114--4124. PMLR, 2019.

\bibitem[Monti et~al.(2019)Monti, Zhang, and Hyvarinen]{monti2019causal}
R.~P. Monti, K.~Zhang, and A.~Hyvarinen.
\newblock Causal discovery with general non-linear relationships using
  non-linear {ICA}.
\newblock \emph{35th Conference on Uncertainty in Artificial Intelligence
  ({UAI})}, 2019.

\bibitem[Noroozi and Favaro(2016)]{noroozi2016unsupervised}
M.~Noroozi and P.~Favaro.
\newblock Unsupervised learning of visual representations by solving jigsaw
  puzzles.
\newblock In \emph{European Conference on Computer Vision ({ECCV})}, pages
  69--84. Springer, 2016.

\bibitem[Oord et~al.(2018)Oord, Li, and Vinyals]{oord2018representation}
A.~v.~d. Oord, Y.~Li, and O.~Vinyals.
\newblock Representation learning with contrastive predictive coding.
\newblock \emph{arXiv:1807.03748}, 2018.

\bibitem[Pearl(2000)]{pearl2000causality}
J.~Pearl.
\newblock \emph{Causality: models, reasoning and inference}, volume~29.
\newblock Springer, 2000.

\bibitem[Poldrack et~al.(2011)Poldrack, Mumford, and
  Nichols]{poldrack2011handbook}
R.~A. Poldrack, J.~A. Mumford, and T.~E. Nichols.
\newblock \emph{Handbook of functional MRI data analysis}.
\newblock Cambridge University Press, 2011.

\bibitem[Poldrack et~al.(2015)Poldrack, Laumann, Koyejo, Gregory, Hover, Chen,
  Gorgolewski, Luci, Joo, Boyd, Hunicke-Smith, Simpson, Caven, Sochat, Shine,
  Gordon, Snyder, Adeyemo, Petersen, Glahn, Reese~Mckay, Curran, G{\"o}ring,
  Carless, Blangero, Dougherty, Leemans, Handwerker, Frick, Marcotte, and
  Mumford]{poldrack2015long}
R.~A. Poldrack, T.~O. Laumann, O.~Koyejo, B.~Gregory, A.~Hover, M.-Y. Chen,
  K.~J. Gorgolewski, J.~Luci, S.~J. Joo, R.~L. Boyd, S.~Hunicke-Smith, Z.~B.
  Simpson, T.~Caven, V.~Sochat, J.~M. Shine, E.~Gordon, A.~Z. Snyder,
  B.~Adeyemo, S.~E. Petersen, D.~C. Glahn, D.~Reese~Mckay, J.~E. Curran,
  H.~H.~H. G{\"o}ring, M.~A. Carless, J.~Blangero, R.~Dougherty, A.~Leemans,
  D.~A. Handwerker, L.~Frick, E.~M. Marcotte, and J.~A. Mumford.
\newblock Long-term neural and physiological phenotyping of a single human.
\newblock \emph{Nature communications}, 6:\penalty0 8885, 2015.

\bibitem[Shimizu et~al.(2006)Shimizu, Hoyer, Hyv{\"a}rinen, and
  Kerminen]{shimizu2006linear}
S.~Shimizu, P.~O. Hoyer, A.~Hyv{\"a}rinen, and A.~Kerminen.
\newblock A linear non-{G}aussian acyclic model for causal discovery.
\newblock \emph{Journal of Machine Learning Research}, 7:\penalty0 2003--2030,
  2006.

\bibitem[Sprekeler et~al.(2014)Sprekeler, Zito, and
  Wiskott]{sprekeler2014extension}
H.~Sprekeler, T.~Zito, and L.~Wiskott.
\newblock An extension of slow feature analysis for nonlinear blind source
  separation.
\newblock \emph{Journal of machine learning research}, 15:\penalty0 921--947,
  2014.

\bibitem[Wasserman(2006)]{wasserman2006all}
L.~Wasserman.
\newblock \emph{All of nonparametric statistics}.
\newblock Springer, 2006.

\bibitem[Wiskott and Sejnowski(2002)]{wiskott2002slow}
L.~Wiskott and T.~Sejnowski.
\newblock Slow feature analysis: Unsupervised learning of invariances.
\newblock \emph{Neural computation}, 14\penalty0 (4):\penalty0 715--770, 2002.

\end{thebibliography}
 \bibliographystyle{abbrvnat} 
\end{document}